\pdfoutput=1
\documentclass[11pt]{article}
\usepackage[final]{acl}
\usepackage{times}
\usepackage{latexsym}
\usepackage[T1]{fontenc}
\usepackage[utf8]{inputenc}
\usepackage{microtype}
\usepackage{inconsolata}
\usepackage{comment}
\usepackage{graphicx}
\usepackage{amsmath}
\usepackage{amssymb}
\usepackage{amsfonts}
\usepackage{array}
\usepackage{multirow}
\usepackage{booktabs}
\usepackage{adjustbox}
\usepackage{subcaption}
\usepackage{stmaryrd}
\usepackage{siunitx}
\usepackage{caption}
\usepackage{url}
\usepackage{bm}
\usepackage{longtable}

\newcommand{\KL}{\mathrm{KL}}
\newcommand\blfootnote[1]{
  \begingroup
  \renewcommand\thefootnote{}\footnote{#1}
  \addtocounter{footnote}{-1}
  \endgroup
}

\title{Establishing a Scale for Kullback--Leibler Divergence\\ in Language Models Across Various Settings}

\author{
{\bf Ryo Kishino}${}^{1}$\qquad 
{\bf Yusuke Takase}${}^{1}$\qquad {\bf Momose Oyama}${}^{1,2}$\\ 
{\bf Hiroaki Yamagiwa}${}^{3}$\thanks{Work done while at Kyoto University.} \qquad {\bf Hidetoshi Shimodaira}${}^{1,2}$\\
${}^{1}$Kyoto University\quad
${}^{2}$RIKEN\quad
${}^{3}$SB Intuitions\\
\texttt{kishino.ryo.32s@st.kyoto-u.ac.jp},\\
\texttt{\{y.takase, oyama.momose\}@sys.i.kyoto-u.ac.jp},\\
\texttt{hiroaki.yamagiwa@sbintuitions.co.jp}, \texttt{shimo@i.kyoto-u.ac.jp}
}

\begin{document}
\maketitle
\begin{abstract}
Log-likelihood vectors define a common space for comparing language models as probability distributions, enabling unified comparisons across heterogeneous settings. We extend this framework to training checkpoints and intermediate layers, and establish a consistent scale for KL divergence across pretraining, model size, random seeds, quantization, fine-tuning, and layers. Analysis of Pythia pretraining trajectories further shows that changes in log-likelihood space, as measured by the scaling behavior of KL divergence, are much smaller than in weight space, resulting in subdiffusive learning trajectories and early stabilization of language-model behavior despite weight drift.
\blfootnote{Our code is available at \url{https://github.com/shimo-lab/modelmap}.}
\end{abstract}

\section{Introduction} \label{sec:intro}

To understand learning dynamics and intermediate-layer representations in language models, it is essential to quantify changes in model behavior and compare them across models.
While traditional analyses rely on weight parameters~\cite{DBLP:journals/nn/ChenQG22, DBLP:journals/neco/KuninSGMTGY24, DBLP:conf/iclr/SinghHHS25, gigante_visualizing_2019}, weight permutation symmetries~\cite{HECHTNIELSEN1990129} and architectural dependencies hinder direct comparisons between models with different learning methods or designs.

\begin{figure}[!t] 
    \centering
    \includegraphics[width=0.95\linewidth]{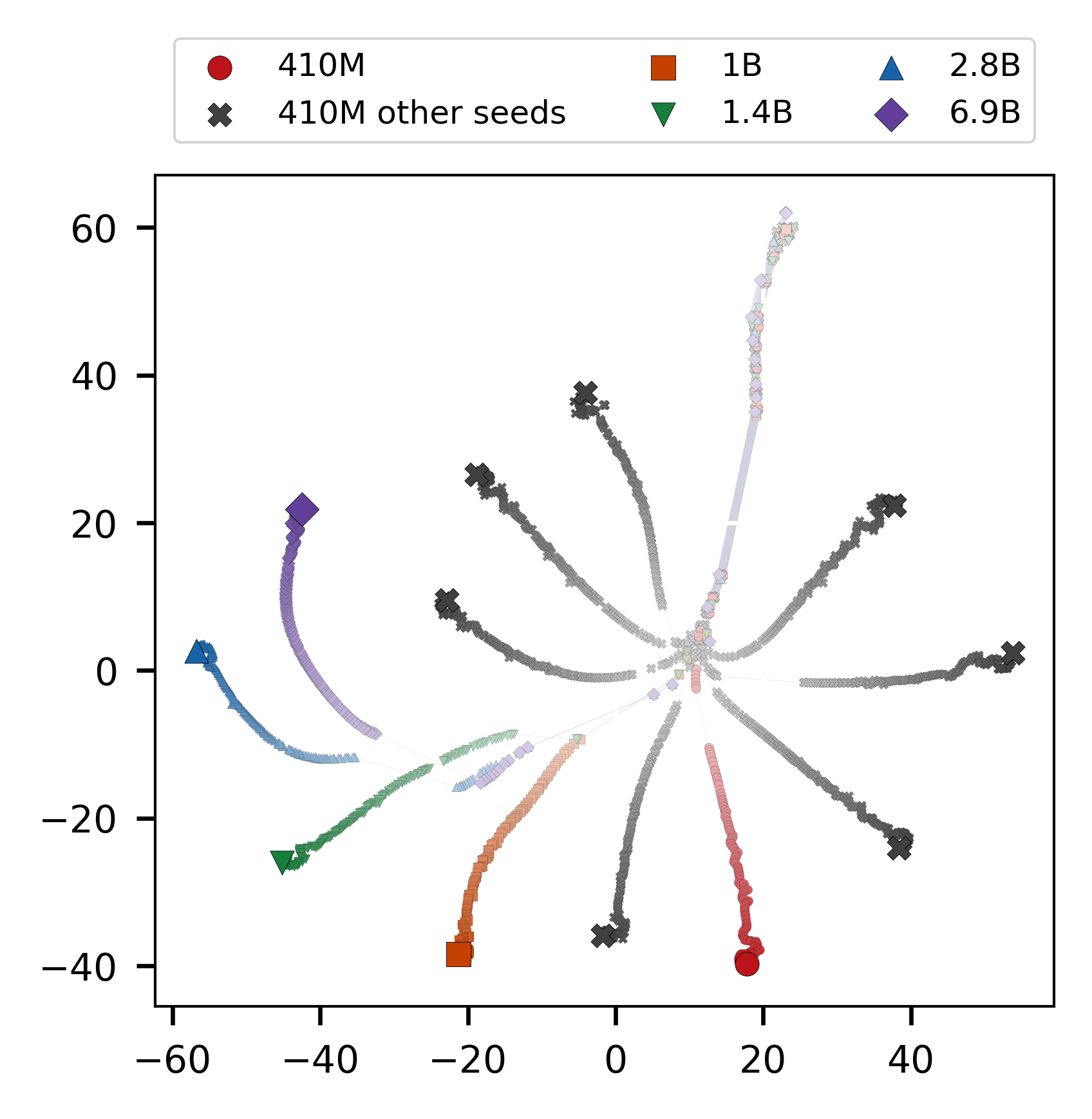}
    \caption{
Visualization of the pretraining trajectories of Pythia models across five model sizes and seven random seeds using t-SNE (perplexity=30) applied to double-centered log-likelihood vectors. Each point represents a model checkpoint, with lighter colors indicating earlier steps. The thickness of the connecting lines reflects the magnitude of KL divergence between successive checkpoints. See Appendix~\ref{app:visual_pretraining} for additional visual analyses.
    }
    \label{fig:pretraining_tsne_std}
    \vspace{-0.3cm}
\end{figure}

\citet{DBLP:conf/acl/OyamaYTS25} introduced the log-likelihood vector for large-scale comparison of language models, representing each model by its probabilities over a set of texts. They showed that models with different architectures can be embedded in the same space. Moreover, since the squared Euclidean distance in this space estimates the Kullback--Leibler (KL) divergence, model comparison can be formulated as a geometric problem.

\begin{figure*}[t]
    \centering
    \includegraphics[width=0.95\linewidth]{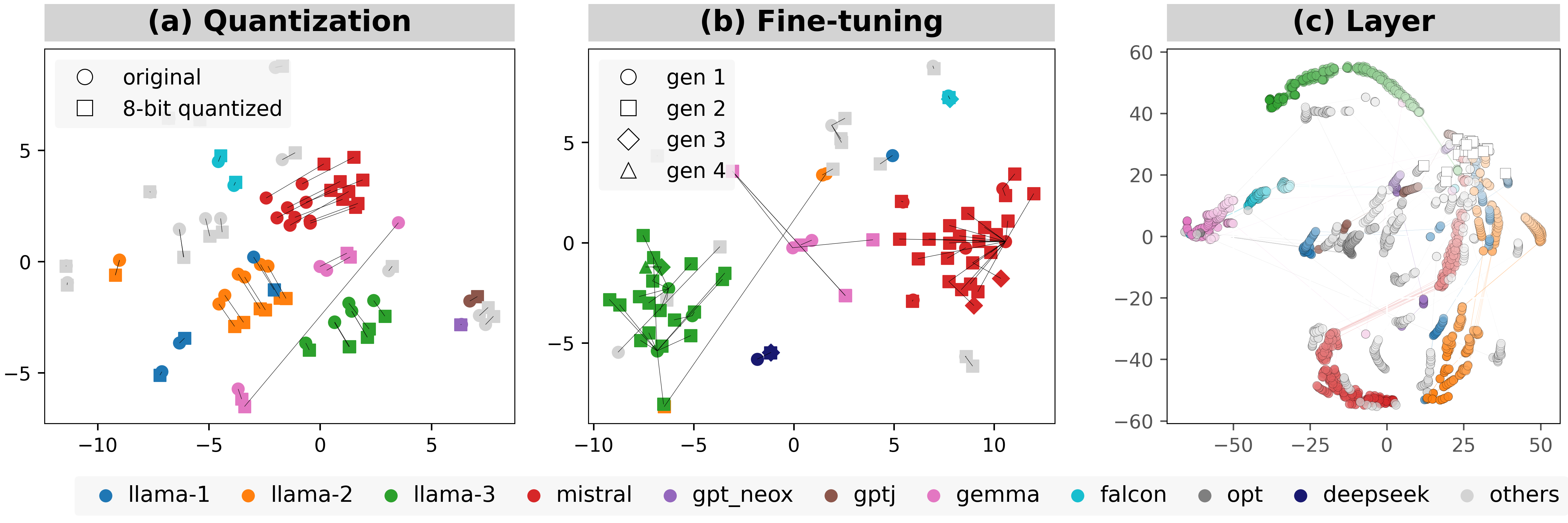}
\caption{
t-SNE visualizations in the space of double-centered log-likelihood vectors.
(a) Original models and their 8-bit quantized models (perplexity=30).
(b) Base models and their fine-tuned models (perplexity=20). Lines represent parent-child relationships, and generation refers to the depth from the root model within each lineage.
(c) Trajectories across layers for 50 models (perplexity=30), obtained by applying the logit lens to each layer. Deeper layers are represented with lighter colors, and the final layer is indicated by a white square.
}
    
    \label{fig:ft_layer_tsne}
    \vspace{-0.3cm}
\end{figure*}

In this study, we demonstrate that log-likelihood vectors provide a consistent representation not only for fully trained models but also for training checkpoints, quantized models, and intermediate layers. We therefore handle diverse models within a single coordinate system (see Figs.~\ref{fig:pretraining_tsne_std} and \ref{fig:ft_layer_tsne}). We establish KL divergence scales across various settings, including checkpoints, model sizes, random seeds, quantization, fine-tuning, and layers as practical metrics for model comparison.

Moreover, analysis of Pythia pretraining trajectories via KL-scaling exponents shows that, despite large changes in weight space, the corresponding changes in log-likelihood space remain remarkably small. This discrepancy appears as subdiffusive behavior in later training, suggesting early stabilization of the model's output distribution despite continued weight drift.

\section{Preliminaries: Model Map and KL} \label{sec:review}

\paragraph{Log-likelihood vector.}

The probability that a language model \( p \) generates a text (i.e., a sequence of tokens) \( x = (y_1, \ldots, y_M) \) is given by $ p(x) = \prod_{m=1}^M p(y_m \mid y_0, \ldots, y_{m-1})$. \citet{DBLP:conf/acl/OyamaYTS25} defined the log-likelihood vector of a model \( p \) over a predefined text set\footnote{\citet{olsson2022context} used token-level log-likelihoods.} \( \{x_1, \ldots, x_N\} \) as
\[
    \bm{\ell} = (\log p(x_1), \ldots, \log p(x_N))^\top \in \mathbb{R}^N,
\]
and showed that it is a useful feature representation for model comparison and downstream performance prediction. This method has also been used in subsequent work such as \citet{harada-etal-2025-massive}.

\paragraph{KL divergence.}
Let \(\{p_i\}_{i=1}^K\) be a set of language models, and \(\bm{L} = (\bm{\ell}_1, \ldots, \bm{\ell}_K)^\top \in \mathbb{R}^{K \times N}\) be the matrix formed by stacking the log-likelihood vectors \(\bm{\ell}_i\) of each model. Define \(\bm{Q} = (\bm{q}_1, \ldots, \bm{q}_K)^\top\) as the matrix obtained by double-centering\footnote{Row-wise centering (over texts) and column-wise centering (over models), respectively, of the log-likelihood matrix.} \(\bm{L}\). Then, the KL divergence between two models \(p_i\) and \(p_j\) can be approximated as
\[
    2\KL(p_i, p_j) \approx \|\bm{q}_i - \bm{q}_j\|^2 / N.
\]
In other words, by associating each model \(p_i\) with a coordinate \(\bm{q}_i \in \mathbb{R}^N\),
KL divergence is approximated by squared Euclidean distance in this space.
We refer to the \(\bm q\)-coordinate system as the model map throughout this paper.
For the interpretation of the KL divergence, the standard error, and the dependence on the text set, see Appendix~\ref{app:KL_SE}.

\paragraph{Model map for various settings.}

Log-likelihood vectors can be computed not only for fully trained language models but also for models during training (Figs.~\ref{fig:pretraining_tsne_std} and \ref{fig:ft_layer_tsne}b) and for quantized models (Fig.~\ref{fig:ft_layer_tsne}a). Moreover, by treating the network up to each layer as a single model via the logit lens~\cite{nostalgebraist2020logitlens}, intermediate layers can also be analyzed (Fig.~\ref{fig:ft_layer_tsne}c). These analyses are difficult to carry out with weight-based approaches~\cite{gigante_visualizing_2019, DBLP:journals/nn/ChenQG22, DBLP:journals/neco/KuninSGMTGY24, DBLP:conf/iclr/SinghHHS25}.

\paragraph{General experimental settings.}
For the text set, we use the same 10,000 texts extracted from the Pile~\cite{arxiv:2101.00027} as in \citet{DBLP:conf/acl/OyamaYTS25}.
The values of KL divergence are normalized by the average text length, $\bar{B} = 972.3188$ bytes, and are reported in units of bits per byte.
Accordingly, the matrix $\bm{Q}$ is divided in advance by $\sqrt{2N\bar{B}\log 2}$, so that the squared Euclidean distance between rows of $\bm{Q}$ directly approximates the KL divergence in bits per byte.

\begin{figure*}[t]
  \centering
  \includegraphics[width=1.0\linewidth]{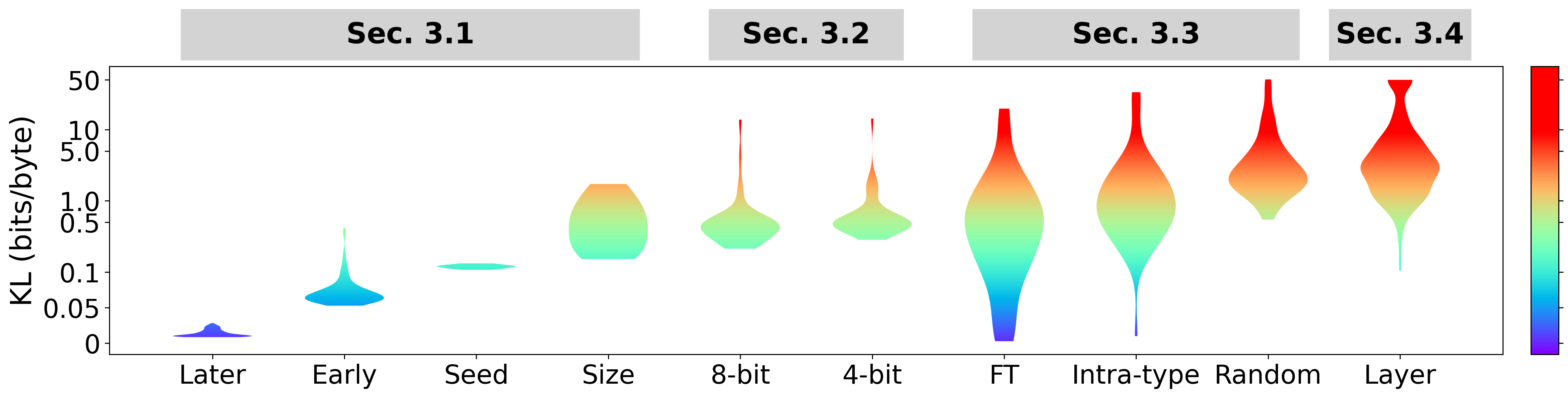}
  \caption{
Scale of KL divergence (bits per byte) across various settings.
The first four entries from the left correspond to pretraining-related comparisons, showing the KL divergence between consecutive saved checkpoints at 1k-step intervals during the last 50k steps of training, at 1k-step intervals during the first 50k steps, across different random seeds at the final checkpoint, and across different model sizes.
The subsequent entries show the KL divergence before and after quantization (8-bit and 4-bit), before and after fine-tuning, between randomly selected pairs within the same model type, between randomly selected pairs across all model types, and between adjacent layers.
The vertical axis is shown on a \texttt{symlog} scale. See Appendix~\ref{app:kl_table} for more details.
}
  \vspace{-0.3cm}
  \label{fig:kl_cheat_sheet}
\end{figure*}

\section{KL Divergence Across Various Settings}
\label{sec:kl_cheat_sheet}

In this section, we measure and summarize KL divergence under a variety of settings, and present the results in Fig.~\ref{fig:kl_cheat_sheet}.
KL divergence is an interpretable quantity on an absolute scale; for example, a value of 0.1 bits/byte corresponds to only 0.1 bit per 8-bit text unit, whereas 10 bits/byte indicates a very large difference, exceeding the 8-bit scale of the text itself\footnote{Since KL divergence is unbounded, it can exceed 8 bits/byte despite the 8-bit representation of text.}.
As a reference point, the effective per-byte entropy of the text set in our experiments is approximately 0.71 bits/byte (see Appendix~\ref{app:KL_SE}).
Our experiments reveal that each setting is associated with a characteristic scale of KL divergence. Even the value of 0.1 bits/byte can be interpreted as a substantial change between consecutive 1k-step checkpoints during training, but as a minor difference when comparing models of different types.

\subsection{Pretraining}
\label{subsec:kl_training}
\paragraph{Settings.}

We use publicly available pretraining checkpoints from the Pythia suite~\cite{arxiv:2304.01373} for model sizes 410M, 1B, 1.4B, 2.8B, and 6.9B. For the 410M model, checkpoints are available for nine random seeds (1--9) released by \citet{wal2025polypythias}. From these, we exclude seeds 3 and 4 due to training loss spikes detected via KL divergence, leaving seven seeds for our experiments. Checkpoints are saved at training steps 0, 1, 2, 4, \dots, 512, 1k, 2k, 3k, \dots, and 143k.
We exclude the 1B checkpoint at step 116k due to anomalous weights and remove texts exhibiting irregular log-likelihood variations across checkpoints. Details of the preprocessing, data filtering, and anomaly detection procedures for pretraining are provided in Appendix~\ref{app:detail_pretraining}.

\paragraph{KL divergence.}
For all model sizes, the KL divergence between consecutive saved checkpoints at 1k-step intervals after the warmup phase is typically 0.05--0.1 bits/byte during the early stage of training, and becomes extremely small in the later stage, ranging from about 0.01 to 0.05 bits/byte.
At the final checkpoint, the KL divergence across different random seeds for the 410M model is around 0.1 bits/byte, while that across different model sizes ranges from 0.15 to 1.7 bits/byte, indicating substantially larger differences.

\subsection{Quantization}\label{subsec:quantization}

\paragraph{Settings.}
We analyze the impact of post-training quantization~\cite{dettmers2022llmint8,arxiv:2305.14314}. 
We use a subset of the 1,018 language models analyzed in \citet{DBLP:conf/acl/OyamaYTS25}, selecting the 50 most downloaded models (Fig.~\ref{fig:ft_layer_tsne}a); see Appendix~\ref{app:quantization} for details on the models used.

\paragraph{KL divergence.}
The median KL divergence before and after 8-bit quantization is 0.44 bits/byte, while it is slightly larger for 4-bit quantization at 0.49 bits/byte.
As shown in Fig.~\ref{fig:kl_cheat_sheet}, the distributions exhibit low variance in both cases, indicating that the degree of degradation induced by quantization is relatively consistent across models.
Furthermore, in Fig.~\ref{fig:ft_layer_tsne}a, the shifts induced by quantization appear to be aligned in both direction and magnitude within each model type, suggesting that quantization acts as a structured perturbation in the log-likelihood space rather than as random noise.

To quantitatively validate this observation, we computed all pairwise cosine similarities among the log-likelihood difference vectors induced by 8-bit quantization within each model type.
The resulting mean similarities were 0.98 for llama-2 (8 models), 0.91 for llama-3 (7 models), and 0.96 for Mistral (9 models). In contrast, when nine models were randomly sampled irrespective of family, the mean cosine similarity was 0.67, averaged over 100 trials.

\subsection{Fine-tuning Effects}
\label{subsec:fine-tuning}

\paragraph{Settings.}

Out of the 1,018 language models analyzed in \citet{DBLP:conf/acl/OyamaYTS25}, we use 66 pairs of fine-tuned and base models (87 models in total; Fig.~\ref{fig:ft_layer_tsne}b). The parent-child relationships are identified using Hugging Face Hub API. See Appendix~\ref{app:detail_finetuning} for details.

\paragraph{KL divergence.}

Fig.~\ref{fig:kl_cheat_sheet} shows that fine-tuning induces a relatively small but non-negligible change within a model type, smaller than the variation among randomly paired models of the same type and much smaller than the variation across model types; the corresponding median KL divergence values are 0.40, 0.95, and 2.2 bits/byte.

\subsection{Across Layers}
\label{subsec:layer}

\paragraph{Settings.}
We investigate model trajectories across layers by using the logit lens~\cite{nostalgebraist2020logitlens} to treat each subnetwork, from the input up to a given layer, as an individual model. This allows us to trace how model behavior changes across layers. We use the same set of 50 models as in Section~\ref{subsec:quantization} (Fig.~\ref{fig:ft_layer_tsne}c).

\paragraph{KL divergence.}
The KL divergence between adjacent layers spans a wide range, with a median of 3.0 bits/byte and a standard deviation of 13 bits/byte. Moreover, within each model type, models follow similar trajectories across layers, as shown in Fig.~\ref{fig:ft_layer_tsne}c, and exhibit similar layerwise profiles of KL divergence between adjacent layers, as shown in Fig.~\ref{fig:layer_kl} in Appendix~\ref{app:detail_layer}.

\section{Scaling Analysis of KL Divergence Along Pretraining Trajectories}
\label{sec:pretraining}

In this section, we examine the scaling behavior of KL divergence over training time along pretraining trajectories of Pythia models, using the same checkpoints as in Section~\ref{subsec:kl_training}.

\subsection{Trajectory on the Log-Likelihood Space} \label{sec:trajectory2d}

\begin{figure}[t]
    \centering
    \includegraphics[width=1.0\linewidth]{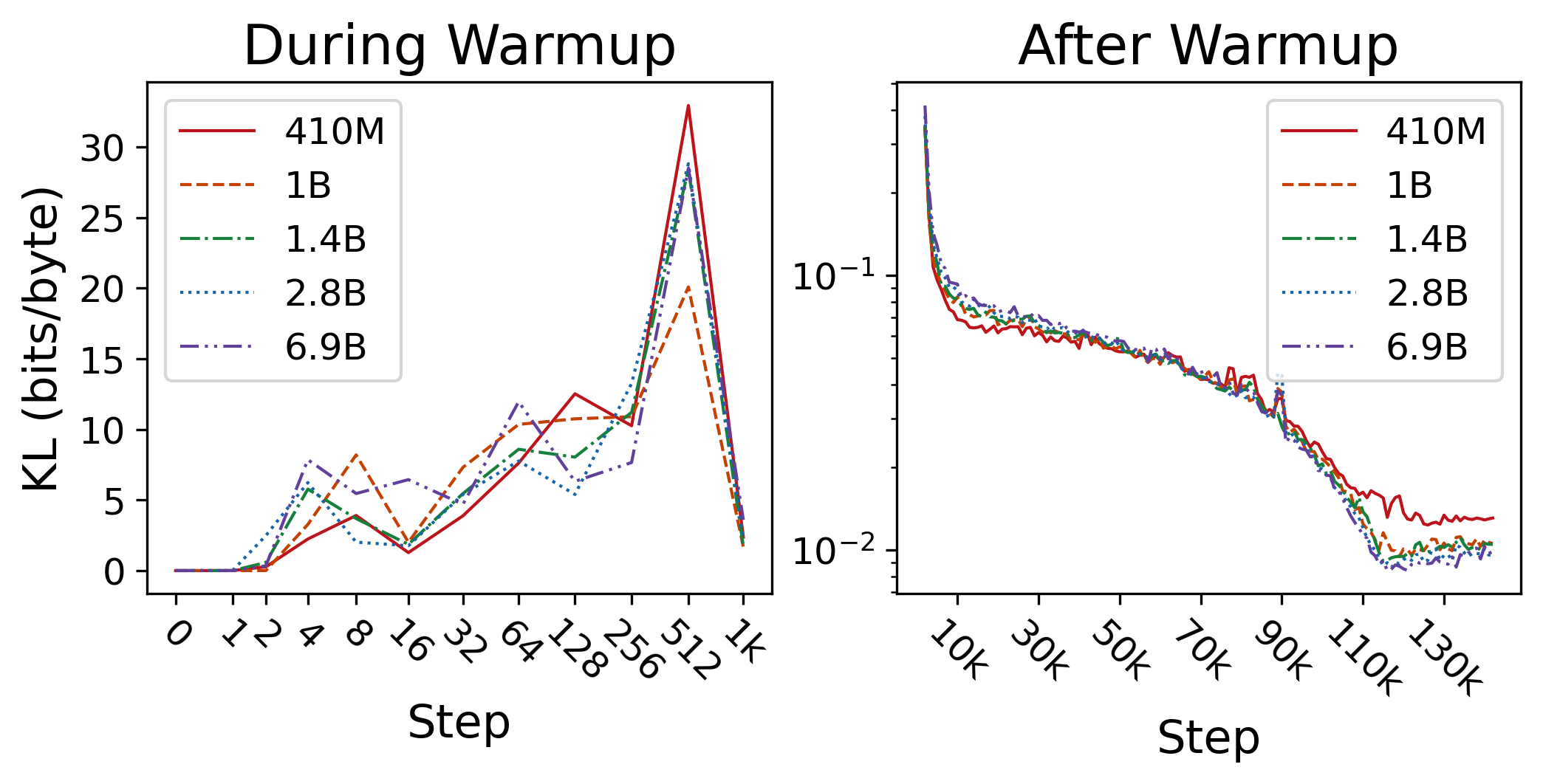}
    \caption{
KL divergence between consecutive saved checkpoints of Pythia during pretraining.
(Left)~Warmup phase with non-uniform checkpoint spacing.
(Right)~Post-warmup phase with checkpoints saved every 1k training steps.
}
\label{fig:trajectory_subsection}
    \vspace{-0.3cm}
\end{figure}

Fig.~\ref{fig:pretraining_tsne_std} visualizes pretraining trajectories of Pythia models across five model sizes and seven random seeds in the log-likelihood space using t-SNE, enabling joint comparisons that are difficult in weight space\footnote{Two-dimensional t-SNE visualizations may exhibit apparent jumps due to dimensionality constraints, whereas three-dimensional t-SNE visualizations alleviate this issue and support continuity of the sampled trajectories at the displayed resolution; see Appendix~\ref{app:visual_pretraining}.}.
Fig.~\ref{fig:trajectory_subsection} shows the KL divergence between consecutive saved checkpoints during pretraining.
After the warmup phase, the KL divergence between checkpoints saved every 1k steps decreases and, after step 10k, remains within a narrow range regardless of model size, whereas substantially larger values are observed during warmup.

\subsection{Comparison with Weight Space} \label{sec:scaling-law}

\paragraph{Anomalous diffusion.}
We compare the stability of training trajectories by estimating diffusion exponents that characterize the power-law scaling of KL divergence.
According to \citet{DBLP:journals/neco/KuninSGMTGY24}, after model performance has converged, the weights diffuse following a power law with exponent $c_{\mathrm w}$.
Specifically, with respect to an initial step $t_0$, the squared Euclidean distance satisfies
\(\|\bm{W}_t - \bm{W}_{t_0}\|^2 \propto |t - t_0|^{c_{\mathrm w}}\).
We show that an analogous power-law relationship also holds for the log-likelihood vectors $\bm{q}_t$:
\(\|\bm{q}_t - \bm{q}_{t_0}\|^2 \propto |t - t_0|^{c_{\mathrm q}}\),
where the left-hand side is proportional to the KL divergence.

\begin{figure}[t]
    \centering
    \includegraphics[width=1.0\linewidth]{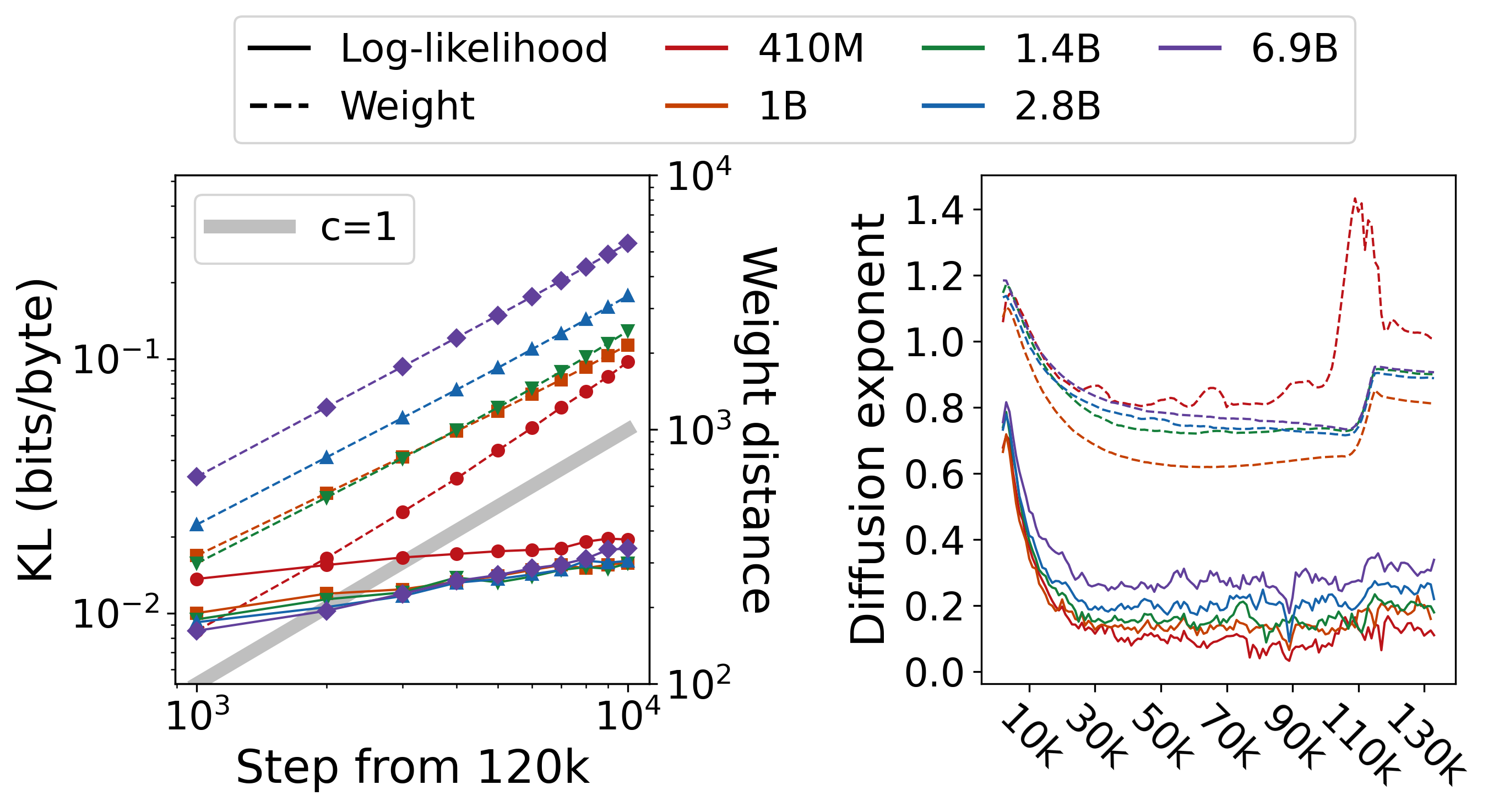}
    \caption{
    (Left) Temporal evolution of the squared Euclidean distance in the weights and log-likelihood vectors from steps 120k to 130k for Pythia. See Table~\ref{tab:diffusion-exponents} for the diffusion exponents in the weight and log-likelihood space for each model size estimated using the least squares method.
    (Right) Diffusion exponent as a function of the starting point $t_0$. 
    The exponent is estimated using the least squares method.
    }
\label{fig:diffusion_subsection}
\end{figure}

For diffusion exponent $c$, Brownian motion corresponds to $c=1$, while $c\neq1$ indicates anomalous diffusion.
In Fig.~\ref{fig:diffusion_subsection}, the weights $\bm{W}$ exhibit Brownian-like diffusion with $c_{\mathrm w} \approx 1$, whereas the diffusion of $\bm{q}$ is substantially suppressed with $c_{\mathrm q} \approx 0.2$. 
The right panel of Fig.~\ref{fig:diffusion_subsection} shows how the exponents $c_{\mathrm w}$ and $c_{\mathrm q}$ vary with the starting step $t_0$.
Compared to weight trajectories, the smaller exponent $c_{\mathrm q}$ indicates that trajectories in log-likelihood space remain confined to a relatively narrow region, suggesting early stabilization of language-model behavior despite continued drift in weight space.

Under a fractional Brownian motion interpretation, these exponents correspond to effective fractal dimensions of $D_{\mathrm w} \approx 2$ and $D_{\mathrm q} \approx 10$, respectively\footnote{If the process is fractional Brownian motion~\cite{mandelbrot1968fractional} with diffusion exponent $c$, the Hurst exponent is $H=c/2$, and the Hausdorff dimension of the trajectory is $D=1/H$~\cite[Theorem 8.4.1]{adler1981geometry}.}.
Thus, the larger effective dimension $D_{\mathrm q}$ suggests greater geometric complexity of the trajectories in log-likelihood space.
See Appendix~\ref{app:discussion_diffusion} for details.

\begin{table}[t]
  \centering
  \small
  \begin{tabular}{lccc|cc}
    \toprule
    Size & $c_{\mathrm{w}}$ & $c_{\mathrm{q}}$ & $\alpha$ & $c_{\exp(\mathrm{q})}$ & Diff\\
    \midrule
    410M & 1.1  & 0.15 & 0.14 & 0.15 & +0.00\\
    1B   & 0.83 & 0.20 & 0.24 & 0.20 & +0.00\\
    1.4B & 0.91 & 0.21 & 0.23 & 0.21 & +0.00\\
    2.8B & 0.90 & 0.26 & 0.29 & 0.26 & +0.00\\
    6.9B & 0.92 & 0.33 & 0.36 & 0.34 & +0.01\\
    \bottomrule
  \end{tabular}
  \caption{
  Diffusion exponents and H\"older exponents for each model size. The checkpoints of Pythia from steps 120k to 130k are used. $\alpha = c_{\mathrm{q}} / c_{\mathrm{w}}$ denotes the H\"older exponent. ``Diff'' denotes $c_{\exp(\mathrm{q})} - c_{\mathrm{q}}$, where $c_{\exp(\mathrm{q})}$ represents the diffusion exponent in the likelihood space.
  }
  \label{tab:diffusion-exponents}
\end{table}

\paragraph{Geometric interpretation via folding.}
The suppressed diffusion exponent $c_{\mathrm q}$ may be understood through the many-to-one structure of the mapping from weight space to log-likelihood space.
Due to permutation symmetries of hidden units, many distinct weight configurations correspond to the same function or output distribution~\cite{fukumizu2000local}.
From a statistical viewpoint, such permutation-induced redundancies are associated with degenerate directions of the Fisher information matrix, giving rise to flat directions in the loss landscape.
As a result, trajectories that are well separated in weight space can be mapped to nearby or identical points in the log-likelihood space.
Viewed through this many-to-one mapping, training trajectories in the log-likelihood space are strongly folded, leading to complex geometric structures that remain confined to a relatively narrow region of the space.

\paragraph{Quantifying folding via H\"older regularity.}
The geometric interpretation based on folding can be further quantified by considering the regularity of the mapping
$f:\bm W\mapsto\bm q(\bm W)$ from weight space to log-likelihood space.
Specifically, a mapping is $\alpha$-H\"{o}lder continuous if distances in the output space are bounded by a constant times the $\alpha$-th power of distances in the input space, with $\alpha=1$ corresponding to Lipschitz continuity and smaller values of $\alpha>0$ indicating increasing non-smoothness.
If $f$ is $\alpha$-H\"{o}lder continuous and satisfies the non-degeneracy condition required to saturate the theoretical bound, then the fractal dimensions of the corresponding trajectories obey
$D_{\mathrm q} = D_{\mathrm w}/\alpha$~\cite{falconer2014fractal}, which implies
\[
\alpha = \frac{D_{\mathrm w}}{D_{\mathrm q}} = \frac{c_{\mathrm q}}{c_{\mathrm w}}.
\]
This relation shows that smaller $\alpha$ corresponds to stronger folding: the weight-space trajectory is mapped to a geometrically more complex trajectory in log-likelihood space while its large-scale displacement is strongly suppressed.
Related discussions of $\alpha$-H\"older regularity and learning dynamics also appear in \citet{ly_optimization_2025}, although the setting there differs from ours.
In our empirical setting, the observed diffusion exponents imply effective dimensions $D_{\mathrm w}\approx2$ and $D_{\mathrm q}\approx10$, yielding an effective H\"{o}lder exponent of $\alpha\approx0.2$.
Table~\ref{tab:diffusion-exponents} also suggests that $\alpha$ tends to increase with model size, raising the possibility that it may reflect underlying properties of trained language models as manifested in the geometry of training trajectories.

\paragraph{Comparison on the likelihood scale.}
A possible concern is that the anomalous diffusion reflected in $c_\mathrm{q}$ might be an artifact of the logarithmic scaling used to define the log-likelihood coordinates.
To address this, we repeated the same analysis on the likelihood scale using exponentiated coordinates, i.e., $\exp(\bm{q})$.
As shown in Table~\ref{tab:diffusion-exponents}, the resulting diffusion exponents $c_{\exp(\mathrm{q})}$ are quantitatively consistent with those obtained from $\bm q$, with differences within $\pm 0.01$ across model sizes.

\section{Conclusion} \label{sec:conclusion}

We measured KL divergence between language models across a wide range of conditions and established a unified and interpretable scale. 
We further examined the scaling behavior of KL divergence along pretraining trajectories.

\section*{Limitations}
\begin{itemize}
    \item
We conduct our experiments using the same set of 10,000 text chunks as those used in \citet{DBLP:conf/acl/OyamaYTS25}, and the measured KL divergence therefore depends on the choice of the text set.
However, we verified that different random subsets of 10,000 text chunks from the same Pile corpus yield highly consistent KL estimates (Appendix~\ref{app:KL_SE}), while the impact of domain shifts across text sets has not been examined.

    \item 
The set of models analyzed in this study is limited and does not fully cover the diversity of existing language models.
In particular, pretraining trajectory analyses are conducted only on the Pythia family, and experiments involving fine-tuning, quantization, and layer-wise analysis rely on selected subsets of models due to data availability and metadata constraints, such as incomplete base-model information in model cards.

    \item
In the layer-wise analysis, the logits obtained from shallow layers via the logit lens contain substantial noise. This issue could be addressed by using the tuned lens~\cite{belrose2023elicitinglatentpredictionstransformers}, an improved version of the logit lens. However, the number of publicly available pretrained tuned lenses is currently limited. We believe this limitation can be overcome in future work by training tuned lenses.

    \item
The checkpoints for Pythia are available only at 1k-step intervals except for the early stages of training. Accordingly, our analysis focuses on trajectory dynamics at scales of 1k steps or larger, while finer-scale behavior below this resolution is not examined. This limitation does not affect our analysis of dynamics at coarser temporal scales.

\item 
The effective H\"older exponent $\alpha$ in our analysis is only a trajectory-based reference value. H\"older regularity of a mapping $f$ is, in principle, defined through local variation in all directions around each point, whereas we observe only the variation of $f$ along training trajectories. Moreover, these trajectories are shaped by optimization of log-likelihood rather than being truly random. Therefore, the estimated $\alpha$ should not be interpreted as a full characterization of the regularity of $f$, but only as an indicator of its behavior along training trajectories.

    \item
While we observe structured behaviors such as subdiffusion in the log-likelihood space, this work represents an initial attempt at trajectory analysis in this space, and our analysis of their underlying causes and implications for learning dynamics remains preliminary and has not yet been quantitatively established.

\end{itemize}

\section*{Acknowledgments}
We thank Luis Iván Hernández Ruíz for helpful discussions.
This work was partially supported by JSPS KAKENHI JP22H05106 and JP23H03355 (to HS), JST CREST JPMJCR21N3 (to HS), JSPS KAKENHI JP25K24366 (to HY), and JST BOOST JPMJBS2407 (to YT and MO).

\bibliography{custom, model_list}

\appendix

\section{Details of Fig.~\ref{fig:kl_cheat_sheet}}
\label{app:kl_table}

In Fig.~\ref{fig:kl_cheat_sheet}, the vertical axis, i.e., the KL divergence values, is plotted using a symmetric logarithmic scale (symlog). The scale is linear up to 0.1 bits/byte and logarithmically scaled beyond that point. For the color bar, values are clipped at 10 bits/byte.

Detailed statistics corresponding to Fig.~\ref{fig:kl_cheat_sheet} are presented in Table~\ref{tab:kl_scale}. In addition, for Pythia-410M, we report the KL divergence between checkpoints separated by 1k steps during pretraining, as well as the KL divergence across different seeds and model sizes at the final checkpoint, in Table~\ref{tab:pretraining_kl}. Furthermore, for the three concrete models, the KL divergence between selected layers is shown in Table~\ref{tab:layer_kl}.

\begin{table*}[t]
\centering
\scriptsize

\begin{subtable}{.99\textwidth}
\centering
\begin{tabular}{lrrrrrrr}
\toprule
 & Median & Mean & SD \\
 \midrule
Later & 0.011 & 0.014 & 0.0050 \\
Early & 0.067 & 0.078 & 0.047 \\
Seed & 0.12 & 0.12 & 0.0071 \\
Size & 0.48 & 0.61 & 0.47 \\
8-bit & 0.44 & 0.80 & 1.9 \\
4-bit & 0.49 & 0.91 & 2.0 \\
FT & 0.40 & 1.5 & 3.6 \\
Intra-type & 0.95 & 2.8 & 6.4 \\
Random & 2.2 & 5.3 & 8.8 \\
Layer & 3.0 & 8.1 & 13 \\
\bottomrule
\end{tabular}
\subcaption{Summary statistics (median, mean, SD) of KL divergence across various settings. This table corresponds to Fig.~\ref{fig:kl_cheat_sheet}.}
\label{tab:kl_scale}
\end{subtable}

\vspace{0.3cm}

\begin{subtable}{.99\textwidth}
  \centering
  \begin{tabular}{l rrr rrrrr}
    \toprule
      & \multicolumn{3}{c}{between checkpoints}
      & \multicolumn{1}{c}{between seeds} & \multicolumn{4}{c}{between model sizes} \\
    \cmidrule(lr){2-4} \cmidrule(lr){5-5} \cmidrule(lr){6-9}
      & 10k$\to$11k & 50k$\to$51k & 100k$\to$101k
      &  avg. & 1B & 1.4B & 2.8B & 6.9B \\
    \midrule
    \multirow{2}{*}{410M}
      & 0.069 & 0.053 & 0.023 &0.11 & 0.28 & 0.52 & 1.0 & 1.7 \\
      & ($\pm$0.0011)
      & ($\pm$0.0010)
      & ($\pm$0.00042)
      &  ($\pm $ 0.0024)
      & ($\pm$0.010)
      & ($\pm$0.018)
      & ($\pm$0.036)
      & ($\pm$0.060) \\
    \bottomrule
  \end{tabular}
  \caption{
  KL divergence for Pythia 410M, measured between checkpoints saved during pretraining, across different random seeds, and across different model sizes. For the inter-seed and inter-size comparisons, the final checkpoints are used. 
  }
  \label{tab:pretraining_kl}
\end{subtable}

\vspace{0.3cm}

\begin{subtable}{.99\textwidth}
  \centering
  \begin{tabular}{lrrr}
    \toprule
    Model & 1$\to$2 & 16$\to$17 & 31$\to$32 \\
    \midrule
    \multirow{2}{*}{Llama-2-7b-hf} 
      & 0.61 & 1.6 & 1.3 \\
      & ($\pm$0.013) & ($\pm$0.063) & ($\pm$0.10) \\
    \multirow{2}{*}{Meta-Llama-3-8B} 
      & 4.0 & 22 & 1.7 \\
      & ($\pm$0.075) & ($\pm$0.78) & ($\pm$0.037) \\
    \multirow{2}{*}{Mistral-7B-v0.3} 
      & 2.4 & 5.7 & 8.6 \\
      & ($\pm$0.10) & ($\pm$0.12) & ($\pm$0.54) \\
    \bottomrule
  \end{tabular}
  \subcaption{
  KL divergence between adjacent layers within specific models, measured using the logit lens.
  }
  \label{tab:layer_kl}
\end{subtable}
\caption{
KL divergence in bits per byte between language models under various conditions. Values in parentheses represent standard errors.
See Appendix~\ref{app:KL_SE} for the formula used to compute the standard errors.
}
\label{tab:kl}
\vspace{-0.3cm}
\end{table*}

\section{Related Work and Further Discussions on Diffusion Exponents} \label{app:discussion_diffusion}

In this appendix, we discuss related work and possible interpretations of the marked discrepancy between the diffusion behavior in weight space and that in log-likelihood space observed in Section~\ref{sec:scaling-law}.

\paragraph{Anomalous diffusion in weight space.}
It has been reported that, during the training of neural networks, trajectories in weight space can exhibit anomalous diffusion, and this phenomenon has recently attracted considerable research interest~\cite{DBLP:journals/nn/ChenQG22,DBLP:journals/neco/KuninSGMTGY24,ly_optimization_2025}.
These studies analyze training dynamics through the temporal scaling of distances in weight space and report deviations from standard Brownian motion.

\paragraph{Diffusion exponents in weight and log-likelihood spaces.}
In Section~\ref{sec:scaling-law}, we estimated the diffusion exponents $c_{\mathrm w}$ and $c_{\mathrm q}$ for trajectories in weight space and log-likelihood space, respectively.
More generally, estimating such scaling exponents from discretely observed data via power-law scaling over a finite range of scales is a classical issue in fractal analysis; see \citet{gneiting2012estimators} for related methodology and for asymptotic and finite-sample assessments under infill asymptotics\footnote{\citet{gneiting2012estimators} consider a real-valued stochastic process $X(t)$, $t\in \mathbb{R}^d$ and estimate $c$ from finite-scale increment statistics, where the diffusion exponent $c$ is defined via the mean squared displacement scaling $\mathbb{E}(|X(t) - X(t_0)|^2) \propto |t-t_0|^c$. They then relate it to the fractal dimension of the graph $\{(t,X(t))\}$ via $D_{\mathrm{graph}} = d+1-c/2$. 
This differs from the effective dimension interpretation used for our one-parameter training trajectories, for which $d=1$ and hence $D=2/c$.}. Our results show that during the early phase of training, both exponents take relatively large values ($c_{\mathrm w}=1.2$ and $c_{\mathrm q}=0.8$), but decrease as training progresses.
In the later stages, the exponent in weight space approaches $c_{\mathrm w}\approx1$, corresponding to Brownian-like diffusion, whereas the exponent in log-likelihood space becomes much smaller ($c_{\mathrm q}\approx0.2\ll1$), which is characteristic of subdiffusion.

In our discussion, fractional Brownian motion is used purely as an interpretive model to relate the observed diffusion exponents to effective fractal dimensions, rather than as a generative model of the training dynamics.
Under this interpretation, a trajectory with diffusion exponent $c$ has effective fractal dimension given by
\[
D=\frac{2}{c},
\]
following the standard Hausdorff-dimension relation for fractional Brownian motion discussed by \citet{adler1981geometry}; see \citet{mandelbrot1968fractional} for the underlying process model.
Accordingly, the representative later-stage values $c_{\mathrm w}\approx1$ and $c_{\mathrm q}\approx0.2$ correspond to $D_{\mathrm w}\approx2$ and $D_{\mathrm q}\approx10$, respectively.

As shown in the right panel of Fig.~\ref{fig:diffusion_subsection} and Table~\ref{tab:diffusion-exponents} in Section~\ref{sec:pretraining}, the estimated diffusion exponent $c_{\mathrm q}$ exhibits systematic variation across model sizes.
In particular, smaller models tend to show lower values ($c_{\mathrm q}\approx0.1\text{--}0.2$), while larger models exhibit higher values ($c_{\mathrm q}\approx0.2\text{--}0.3$).
While we report $c_{\mathrm q}\approx0.2$ as a representative value capturing the typical subdiffusive behavior in log-likelihood space, a more detailed analysis of model-size dependence is beyond the scope of this paper.

\paragraph{Interpretation via loss landscape and flat minima.}
In our experimental setting, texts are sampled from a corpus that closely approximates the training data.
Consequently, each component of the log-likelihood vector corresponds to a term contributing to the cross-entropy loss, and the negative average of these components serves as a reasonable empirical estimate of the expected loss.
The marked contrast between the near-Brownian diffusion in weight space and the strongly suppressed diffusion in log-likelihood space indicates that, even as model weights continue to change, the output probability distributions remain relatively stable.
This observation is consistent with the hypothesis that stochastic gradient descent tends to favor flat regions of the loss landscape~\cite{keskar2017on}.

\paragraph{Geometric interpretation via folding.}
This optimization-based view is closely related to a more structural interpretation arising from the intrinsic redundancies of multilayer neural networks.
Due to permutation symmetries of hidden units, many distinct weight configurations correspond to the same function or output distribution~\cite{fukumizu2000local}.
From a statistical viewpoint, such permutation-induced redundancies are associated with degenerate directions of the Fisher information matrix, giving rise to flat directions in the loss landscape.
As a result, trajectories that are well separated in weight space can be mapped to nearby or identical points in the log-likelihood space.
Viewed through this many-to-one mapping, training trajectories in the log-likelihood space are strongly folded, leading to complex geometric structures that remain confined to a relatively narrow region of the space.

\paragraph{Quantifying folding via H\"older regularity.}
The geometric interpretation based on folding can be further quantified by considering the regularity of the mapping
$f:\bm W\mapsto\bm q(\bm W)$ from weight space to log-likelihood space.
Specifically, recall that a mapping is $\alpha$-H\"{o}lder continuous if distances in the output space are bounded by a constant times the $\alpha$-th power of distances in the input space, with $\alpha=1$ corresponding to Lipschitz continuity and smaller values of $\alpha>0$ indicating increasing non-smoothness.
If $f$ is $\alpha$-H\"older continuous, then the fractal dimensions of the corresponding trajectories obey
\[
D_{\mathrm q} \le \frac{D_{\mathrm w}}{\alpha}
\]
\cite{falconer2014fractal}. If $f$ additionally satisfies the non-degeneracy condition required to saturate this bound, then the corresponding effective H\"older exponent along the training trajectory is
\[
\alpha = \frac{D_{\mathrm w}}{D_{\mathrm q}} = \frac{c_{\mathrm q}}{c_{\mathrm w}}.
\]
This relation shows that as $\alpha$ decreases, the mapping can strongly fold the weight-space trajectory into the log-likelihood space. As a result, the resulting trajectory becomes geometrically more complex and has a higher effective fractal dimension, while its large-scale displacement remains strongly suppressed.
In our empirical setting, the observed diffusion exponents imply effective dimensions $D_{\mathrm w}\approx2$ and $D_{\mathrm q}\approx10$, which in turn yield an effective H\"{o}lder exponent of $\alpha\approx0.2$.

\paragraph{Interpretation of the effective H\"older exponent.}
The effective H\"older exponent $\alpha = c_{\mathrm q}/c_{\mathrm w}$, computed from the trajectory starting at $t_0$, is used as a trajectory-based proxy for the pointwise H\"older exponent $\alpha(\bm W_{t_0})$ at $\bm W_{t_0}$, in the sense that it reflects the effective roughness of $f$ along the observed training trajectory rather than the actual pointwise regularity defined using all nearby perturbation directions around $\bm W_{t_0}$\footnote{Along the training trajectory starting at $t_0$, we observe the power-law relations
$\|\bm W_t-\bm W_{t_0}\|^2 \propto |t-t_0|^{c_{\mathrm w}}$
and
$\|f(\bm W_t)-f(\bm W_{t_0})\|^2 \propto |t-t_0|^{c_{\mathrm q}}$.
Eliminating $|t-t_0|$ from these two relations yields
$\|f(\bm W_t)-f(\bm W_{t_0})\| \propto \|\bm W_t-\bm W_{t_0}\|^{c_{\mathrm q}/c_{\mathrm w}}$.
This heuristically motivates interpreting $c_{\mathrm q}/c_{\mathrm w}$ as a trajectory-based effective H\"older exponent of $f$ at $\bm W_{t_0}$.}.
Estimating the local regularity of $f$ at $\bm W_{t_0}$ along a fixed direction $\bm\delta$ from the local scaling
$\|f(\bm W_{t_0}+\epsilon\bm\delta)-f(\bm W_{t_0})\| \propto |\epsilon|^{\alpha(\bm W_{t_0}, \bm\delta)}$, where $\epsilon\in\mathbb{R}$, can in principle yield an even larger directional H\"older exponent $\alpha(\bm W_{t_0}, \bm\delta)$ than the trajectory-based estimate reported here.
This does not contradict the present interpretation, because H\"older regularity need not be isotropic, and the pointwise H\"older exponent captures the roughest local behavior, in the sense that $\alpha(\bm W_{t_0}) \le \alpha(\bm W_{t_0}, \bm\delta)$ for any nonzero $\bm\delta$.
Even when $\bm\delta = \bm W_{t_1} - \bm W_{t_0}$, $\alpha(\bm W_{t_0}, \bm\delta)$ can still be larger than the trajectory-based estimate.

\paragraph{A perspective from embedding theory.}
In our analysis, we visualize trajectories in the $N$-dimensional log-likelihood space $\mathbb{R}^N$ in $n=2$ or $n=3$ dimensions via t-SNE. 
This approach naturally raises the question of when such low-dimensional representations remain one-to-one and when self-intersections become unavoidable. According to the theory of \textit{embedology} established by \citet{sauer1991embedology}, for a compact set $A \subset \mathbb{R}^N$ with box-counting dimension\footnote{
Although the Hausdorff dimension and box-counting dimension can differ in a strict mathematical sense, they are expected to be equal in the context of the dynamical systems analyzed here. Thus, we use them interchangeably throughout this heuristic analysis.} $d$, almost every $C^1$ map $F: \mathbb{R}^N \to \mathbb{R}^n$ is one-to-one on $A$ provided $n > 2d$. This property is established in the sense of \textit{prevalence} (i.e., a probability-one notion under finite-dimensional perturbations of the map). Conversely, for $n \le 2d$ and for each fixed $\delta > 0$, the $\delta$-distant self-intersection set has a lower box-counting dimension of at most $2d - n$.

For the practically important case $n=2$, we have $2d-n = 2d-2 \ge 0$ for any $d \ge 1$. Thus, generic injectivity is no longer guaranteed, and self-intersections cannot in general be ruled out. In particular, when $d=1$, the self-intersection set has dimension at most $0$, corresponding heuristically to isolated intersection points. For $n=3$, we have $2d-n = 2d-3$. Hence, if $d \ge 1.5$, self-intersections may still remain, whereas if $d<1.5$, the condition $n>2d$ holds, and the set can generically be represented in three dimensions by a one-to-one map without self-intersections.

We return to this viewpoint in Appendix~\ref{app:tsne_embedology}, where we discuss its heuristic implications for the two- and three-dimensional t-SNE visualizations of the Pythia pretraining trajectories.

\paragraph{A related perspective on H\"older regularity.}
\citet{ly_optimization_2025} also relate anomalous diffusion during training to H\"older regularity.
In their framework, the H\"older exponent is introduced through the loss landscape, which in our setting corresponds not to the full log-likelihood vector but to its one-dimensional component along the $\bm 1$-direction, i.e., to the mean log-likelihood up to a constant scaling.
From this perspective, their analysis may be viewed as emphasizing a one-dimensional aspect of the geometry induced by the log-likelihood vector. Our analysis extends this viewpoint by considering training trajectories in the full log-likelihood space, rather than only the averaged direction. This makes it possible to compare the scaling behavior in weight space and in log-likelihood space, and thereby to quantify how the mapping from weights to model behavior contracts and folds trajectories. 
Another difference lies in the notion of dimension being considered. \citet{ly_optimization_2025} discuss the fractal dimension of the graph of the loss landscape\footnote{In their fractional Brownian surface model, the fractal dimension is $D_{\bar\ell} = \dim\bm{W}+1- \alpha_{\bar\ell}$ with $\alpha_{\bar\ell}$ being the H\"older exponent of the loss landscape.}, which in our setup corresponds to $\{(\bm{W}, \bar\ell(\bm{W}))\}$. By contrast, our $D_{\mathrm w}$ and $D_{\mathrm q}$ are the effective fractal dimensions of the trajectories themselves, namely $\{\bm{W}_t\}$ in weight space and $\{\bm{q}_t\}$ in log-likelihood space, respectively.
Thus, while both studies relate training dynamics to H\"older-type regularity and anomalous diffusion, our formulation adds a geometric comparison between parameter space and behavior space, providing a complementary perspective on how training trajectories are transformed when viewed through model behavior.

\paragraph{An illustrative example of folding.}
Although the mapping $f:\bm W\mapsto\bm q(\bm W)$ is explicitly defined once the neural network architecture and the evaluation text set are fixed, its geometric regularity and folding structure are not directly accessible in closed form.

To illustrate this folding mechanism more concretely, we consider an example based on generalized Takagi--Landsberg functions.
These functions are defined as multiscale summations of sawtooth functions $S(x)=\text{dist}(x,\mathbb{Z})$:
\[
f(\bm{W}) = \sum_{k=1}^{\infty} \mathbf{a}_k \lambda^{-k\alpha}
S\!\left(\lambda^k \mathbf{b}_k^\top \bm{W}\right),
\]
where $\lambda>1$ is a constant and $\mathbf{a}_k$ and $\mathbf{b}_k$ are vectors with the same dimensions as $\bm{q}$ and $\bm{W}$, respectively.
This class of functions is known to be $\alpha$-H\"older continuous for any $0<\alpha<1$ under mild boundedness and non-degeneracy conditions on $\{\mathbf a_k,\mathbf b_k\}$~\cite{shidfar1986continuity,allaart2011takagi}.
Such constructions provide a simple geometric illustration of how the many-to-one and redundant parameterization of deep neural networks can give rise to highly folded trajectories in the log-likelihood space.
In contrast to this construction, previous studies have focused on folding phenomena in input--output mappings of deep networks, where depth induces increasingly fine sawtooth partitions of the input space~\cite{telgarsky2016benefits,telgarsky2017neural}.

\section{On the Estimation of the KL Divergence} \label{app:KL_SE}
\paragraph{Interpretation of KL divergence via text entropy.}

As a reference for interpreting KL divergence values, we estimate the entropy of the text from log-likelihood statistics, as described below.
In our setting, this estimated text entropy is approximately 0.71 bits/byte; for example, a KL divergence of 0.1 bits/byte corresponds to a value slightly over 10\% of the text entropy.

In this work, we use the minimum value of the negative mean log-likelihood across models as an estimate of the text entropy.
This is because, given a model $p_i$, a text set $\{x_1,\ldots,x_N\}$, the log-likelihood $\ell_i(x_s)$, and the true data-generating distribution $p_0$, the following approximation holds:
\begin{align*}
    -\frac{1}{N} & \sum_{s=1}^N \ell_i(x_s) 
    \approx \mathbb{E}_{x\sim p_0}[-\log p_i(x)]\\
    &= H(p_0)  + \mathrm{KL}(p_0, p_i)\\
& \ge H(p_0) ,
\end{align*}
where $H(p_0) = \mathbb{E}_{x\sim p_0}[-\log p_0(x)]$ is the text entropy.
 Therefore, the minimum of the negative mean log-likelihood across models 
 provides an empirical upper bound on the text entropy $H(p_0)$ and may serve as a rough approximation when at least one model is close to $p_0$.

\paragraph{Standard error of the estimated KL divergence.}
Let $\bm{q}_i$ and $\bm{q}_j$ denote the $\bm{q}$ coordinates of models $p_i$ and $p_j$, respectively. While in the main text of this paper we use $\bm{q}$ coordinates rescaled by $1/\sqrt{2N\bar B \log 2}$ so that squared Euclidean distances approximate KL divergence in bits per byte, we do not apply this rescaling in this appendix.
Accordingly, the KL divergence between the models is estimated as
$\KL(p_i, p_j) \approx \frac{1}{2N} \|\bm{q}_i - \bm{q}_j\|^2 = \frac{1}{2N} \sum_{k=1}^N (Q_{ik} - Q_{jk})^2,$
which corresponds to the sample mean of $\{ (Q_{ik} - Q_{jk})^2/2 \}_{k=1}^N$.
Assuming that the effects of double-centering in $\bm{Q}$ are negligible, these terms can be treated as approximately independent. When $N$ is sufficiently large, the variance of the estimated KL divergence is given by
\begin{align*}
    &\mathrm{Var}(\KL(p_i, p_j)) \\
    &\approx \frac{1}{N} \left\{ \frac{1}{4N} \sum_{k=1}^N (Q_{ik} - Q_{jk})^4\right. \\
    &\qquad \left. - \frac{1}{4N^2} \|\bm{q}_i - \bm{q}_j\|^4 \right\} \\
    &\approx \frac{1}{4N^2} \sum_{k=1}^N (Q_{ik} - Q_{jk})^4 - \frac{1}{N} \KL(p_i, p_j)^2.
\end{align*}
Therefore, the standard error of the estimator $\KL(p_i, p_j)$ can be approximated as
\begin{align*}
    &\mathrm{SE}(\KL(p_i, p_j)) \\
    &\approx \sqrt{ \frac{1}{4N^2} \sum_{k=1}^N (Q_{ik} - Q_{jk})^4 - \frac{1}{N} \KL(p_i, p_j)^2 }.
\end{align*}

The standard error estimated in nats above is converted to bits per byte by dividing by $\bar B \log 2$.
Equivalently, when using $\bm{q}$ coordinates rescaled by $1/\sqrt{2N\bar B \log 2}$ as in the main text, the standard error of KL divergence in bits per byte is given by
\begin{align*}
    &\mathrm{SE}(\widetilde \KL(p_i, p_j)) \\
    &\approx \sqrt{ \sum_{k=1}^N (\tilde Q_{ik} - \tilde Q_{jk})^4 - \frac{1}{N} \widetilde \KL(p_i, p_j)^2 },
\end{align*}
where $\tilde Q = Q/\,\sqrt{2N\bar B\log 2} $ and
\[
\widetilde \KL(p_i,p_j)
=
\frac{\KL(p_i,p_j)}{\bar B \log 2}
=
\sum_{k=1}^N (\tilde Q_{ik}-\tilde Q_{jk})^2.
\]

\paragraph{On the robustness to the sampling of texts.}
As an additional experiment, we computed the log-likelihood vectors using a new set of 10,000 texts randomly sampled from the 5,703,791 texts in a subset of the Pile\footnote{The full Pile corpus, when segmented into 1,024-byte text chunks, yields a total of 5,703,791 texts~\cite{DBLP:conf/acl/OyamaYTS25}.}, using the Pythia-410M model. We then computed the pairwise KL divergence between all checkpoints after warmup, and found that the Pearson correlation coefficient between the KL values obtained using the original text set and those obtained using the new text set was 0.99. This indicates that the impact of text sampling on the KL values is limited. 

One advantage of using log-likelihood vectors is that the resolution for each domain can be adjusted by the amount of text from that domain included in the dataset. If a specific domain is fixed, it effectively corresponds to projecting the trajectory onto a certain aspect, and thus the shape of the trajectory is expected to change.

\section{Additional Visual Analyses of Pretraining Trajectories}
\label{app:visual_pretraining}

This appendix provides supplementary visual analyses and interpretation of pretraining trajectories.
Appendix~\ref{app:pretraining_tsne_3d} presents three-dimensional t-SNE visualizations in the log-likelihood space, and Appendix~\ref{app:tsne_embedology} discusses their embedding-theoretic interpretation, including possible sources of apparent discontinuities and oscillatory artifacts.
Appendix~\ref{app:pretraining_pca} then uses PCA to examine the global structure of the trajectories in the log-likelihood space.
In contrast, Appendix~\ref{app:weight_space_visualization} presents trajectory visualizations in the weight space, highlighting fundamental limitations of joint visualization across models.

\subsection{Three-Dimensional t-SNE Visualizations}
\label{app:pretraining_tsne_3d}

\begin{figure}[t]
    \begin{minipage}{0.99\linewidth}
        \centering
        \includegraphics[width=0.9\linewidth]{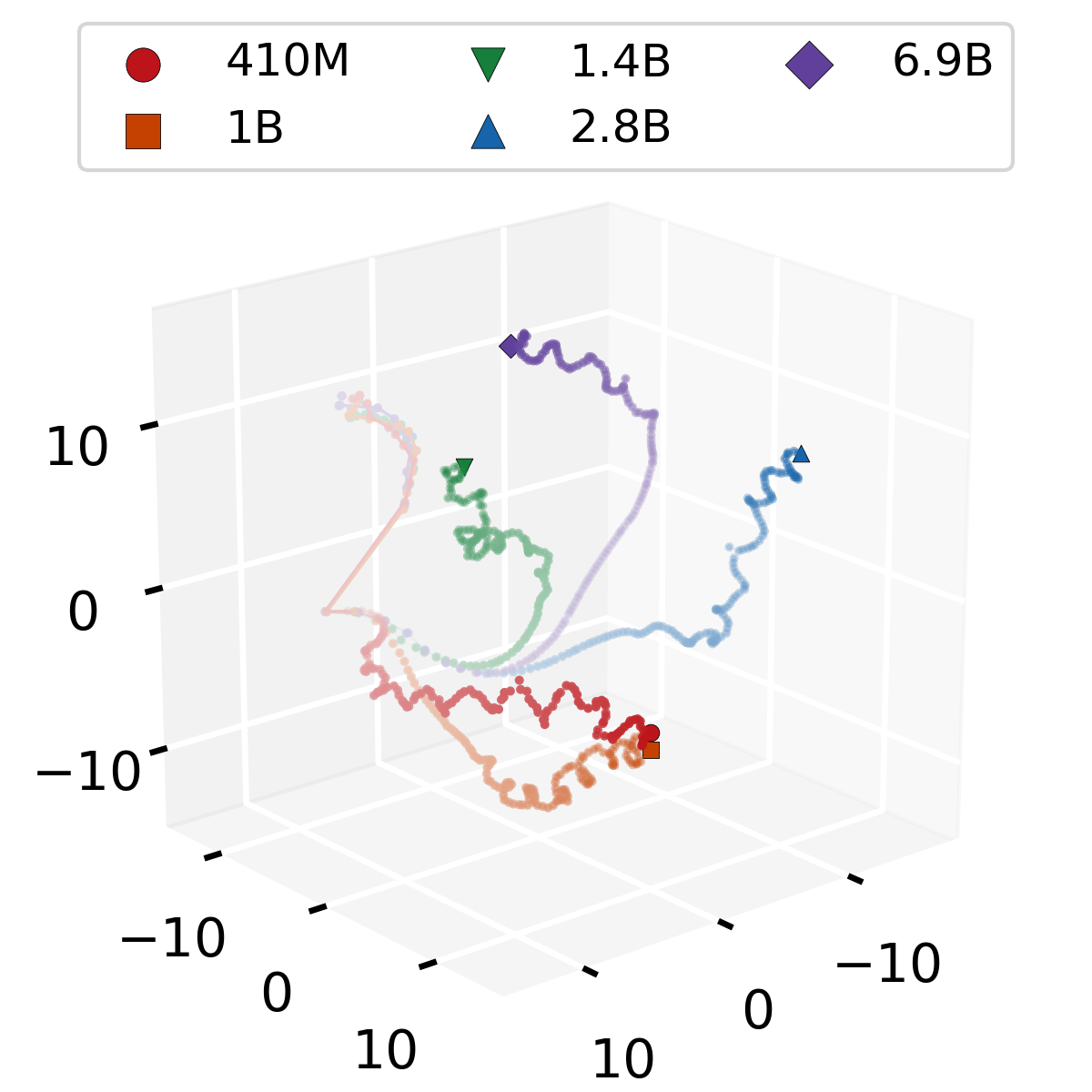}\\
        \subcaption{
        Visualization of pretraining trajectories for five Pythia model sizes using 3D t-SNE (perplexity = 30).
        This figure corresponds to a three-dimensional version of Fig.~\ref{fig:pretraining_tsne_std}, excluding additional random seeds of the 410M model.
        }
        \label{fig:pretraining_3d_tsne}
    \end{minipage}
    \vspace{0.3cm}
    \begin{minipage}{0.99\linewidth}
        \centering
        \includegraphics[width=0.9\linewidth]{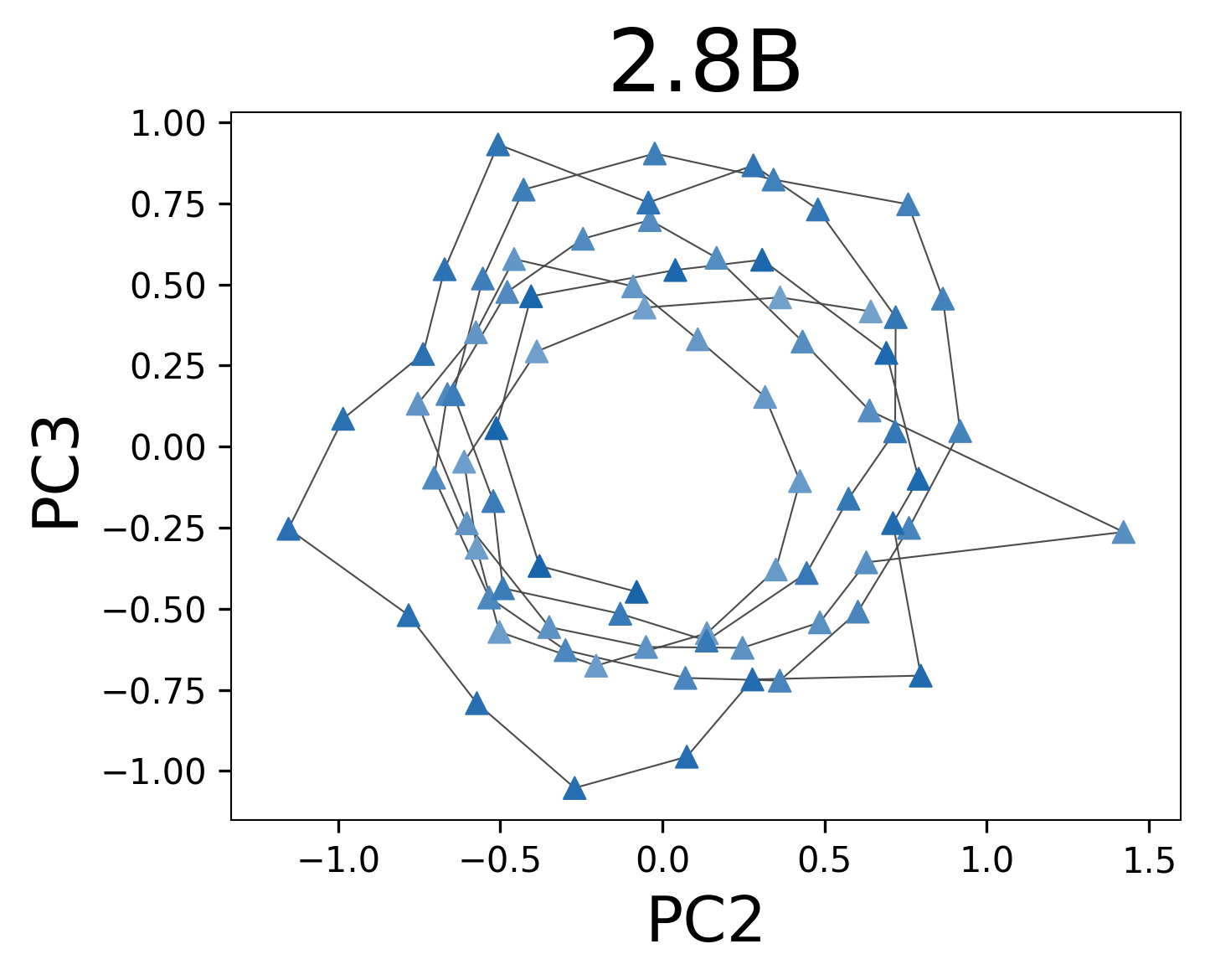}
        \subcaption{
        PC2 versus PC3 for the later portion of the Pythia-2.8B training trajectory after applying 3D t-SNE.
        }
        \label{fig:pretraining_spiral_2.8b}
    \end{minipage}
    \caption{
    Pythia training trajectories visualized using 3D t-SNE.
    }
    \label{fig:pretraining_3d_artifact}
\end{figure}

\begin{figure}[t]
    \centering
    \includegraphics[width=0.8\linewidth]{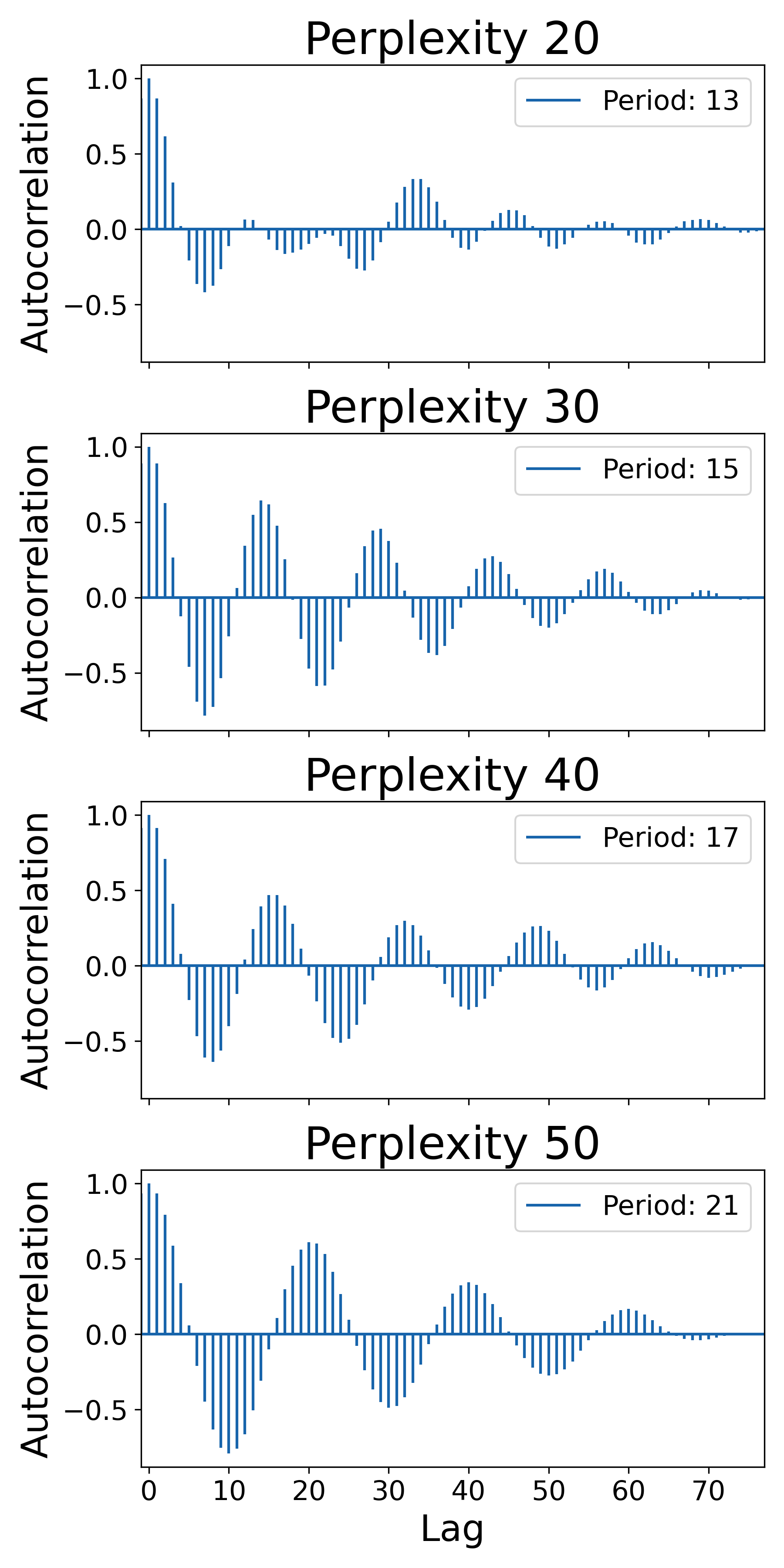}
    \caption{
    Autocorrelation functions of PC3 for the later portion of the Pythia-2.8B training trajectory (from step 67k onward) after applying 3D t-SNE, shown for different perplexity values.}
    \label{fig:pretraining_autocorr_perplexity}
\end{figure}

This subsection presents three-dimensional t-SNE visualizations of the pretraining trajectories discussed in Section~\ref{sec:pretraining}.
In the main text, Fig.~\ref{fig:pretraining_tsne_std} shows two-dimensional t-SNE visualizations of pretraining trajectories for Pythia models across multiple model sizes and random seeds.
In two dimensions, apparent discontinuities may arise near regions where the images of different trajectories overlap.
This phenomenon should be understood as a limitation of low-dimensional visualization rather than as evidence of discontinuity of the underlying trajectories.
A heuristic interpretation of this phenomenon from embedding theory is given in Appendix~\ref{app:tsne_embedology}.

As shown in Fig.~\ref{fig:pretraining_3d_artifact}, this issue is alleviated in the three-dimensional visualization.
Since the visualization is based on finitely many checkpoints sampled along training, the curves shown here should be understood as coarse approximations to the underlying continuous trajectories.
In the 3D t-SNE representation, the sampled training trajectories appear as continuous curves without obvious jumps or self-intersections.
For clarity, we exclude additional seeds of the Pythia-410M model and visualize the trajectories of the five model sizes simultaneously in three dimensions, as shown in Fig.~\ref{fig:pretraining_3d_tsne}.

Figure~\ref{fig:pretraining_spiral_2.8b} further visualizes the later portion of the Pythia-2.8B trajectory by plotting PC2 versus PC3 after applying 3D t-SNE.
In this representation, the trajectory appears to exhibit a spiral-like structure.
To analyze this phenomenon, Fig.~\ref{fig:pretraining_autocorr_perplexity} shows the autocorrelation function of PC3 for the later portion of the trajectory, computed under different t-SNE perplexity settings (20, 30, 40, and 50).
The spiral period is estimated as the training step at which the autocorrelation function attains its maximum between the second and third zero-crossings.
As the perplexity increases, the estimated period also increases, indicating that the apparent spiral structure depends on the neighborhood size used in the embedding.

This spiral-like appearance should therefore be interpreted primarily as a visualization artifact rather than as an intrinsic property of the training dynamics.
At the same time, its emergence may reflect the difficulty of representing a highly folded trajectory in only three dimensions, which is consistent with the subdiffusive scaling $\|q_t-q_{t_0}\|^2 \propto |t-t_0|^{c_q}$ and the large effective fractal dimension inferred from it.

\subsection{Embedding-Theoretic Interpretation of t-SNE Trajectories}
\label{app:tsne_embedology}

We now apply the embedding-theoretic viewpoint discussed in Appendix~B to the pretraining trajectories of Pythia. The actual two- and three-dimensional t-SNE visualizations are given in Section~\ref{sec:trajectory2d} and Appendix~\ref{app:pretraining_tsne_3d}, respectively. If the continuous trajectory in log-likelihood space is interpreted heuristically as a fractal-like compact set with effective dimension $d \approx D_q$, then our estimate $D_q \approx 10$ places it well within the regime $d \ge 1.5$. From this perspective, even a three-dimensional representation is generally not expected to remain one-to-one, and residual folding or self-overlap is therefore not surprising. The oscillatory artifacts observed in Fig.~\ref{fig:pretraining_3d_artifact} may reflect this geometric difficulty of representing a highly folded trajectory in only three dimensions.

At the same time, the actual visualization is constructed from finitely many checkpoints sampled along training. More generally, fractal or roughness-related quantities are defined through limiting behavior at small scales, whereas actual observations are available only over a finite range of scales \citep{gneiting2012estimators}. When these sampled points are connected in temporal order, they form a polygonal chain, which is a one-dimensional object in a coarse geometric sense. Under this approximation, three dimensions become the first case in which a generic one-to-one representation is possible, whereas two dimensions remain a critical case in which apparent overlaps may persist. Although this argument remains heuristic, since t-SNE is not itself a generic $C^1$ map of the type assumed in the embedding theorem and the sampled trajectories are only finite approximations to the underlying continuous path, it is nevertheless consistent with our empirical observations: in the three-dimensional t-SNE visualization, no obvious self-intersections are observed among the sampled trajectories, whereas in the two-dimensional visualization of Fig.~\ref{fig:pretraining_tsne_std}, apparent overlaps can manifest themselves as trajectory jumps.

\subsection{PCA-Based Visualizations of Pretraining Trajectories}
\label{app:pretraining_pca}

\begin{figure}[t] 
    \centering
    \includegraphics[width=1.0\linewidth]{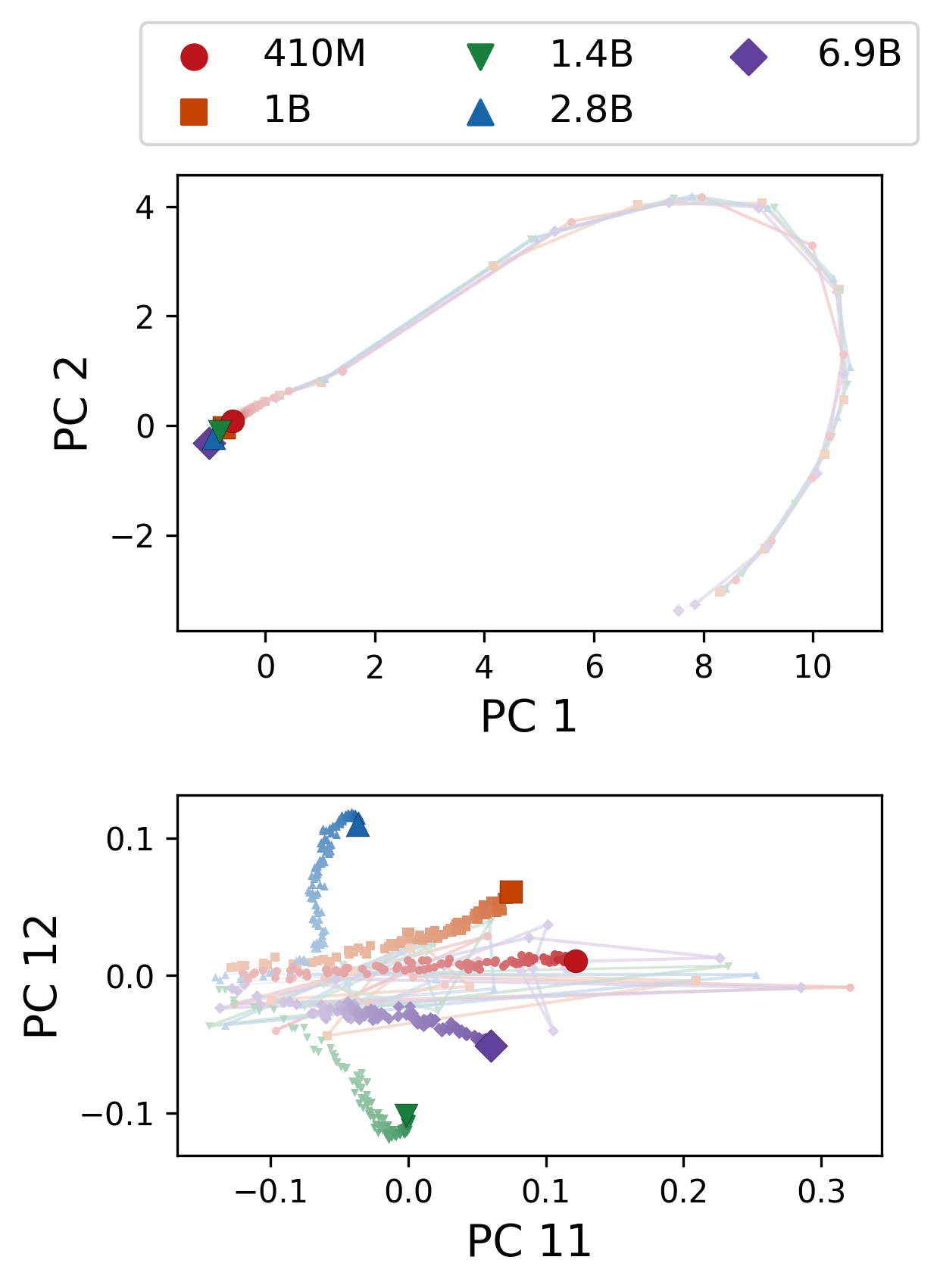}
    \caption{
    PCA visualization of the training trajectories of Pythia models. (Top) PC1 vs. PC2. (Bottom) PC11 vs. PC12.
    }
    \label{fig:pretraining_pca}
\end{figure}

\begin{figure}[t]
    \centering
    \includegraphics[width=0.9\linewidth]{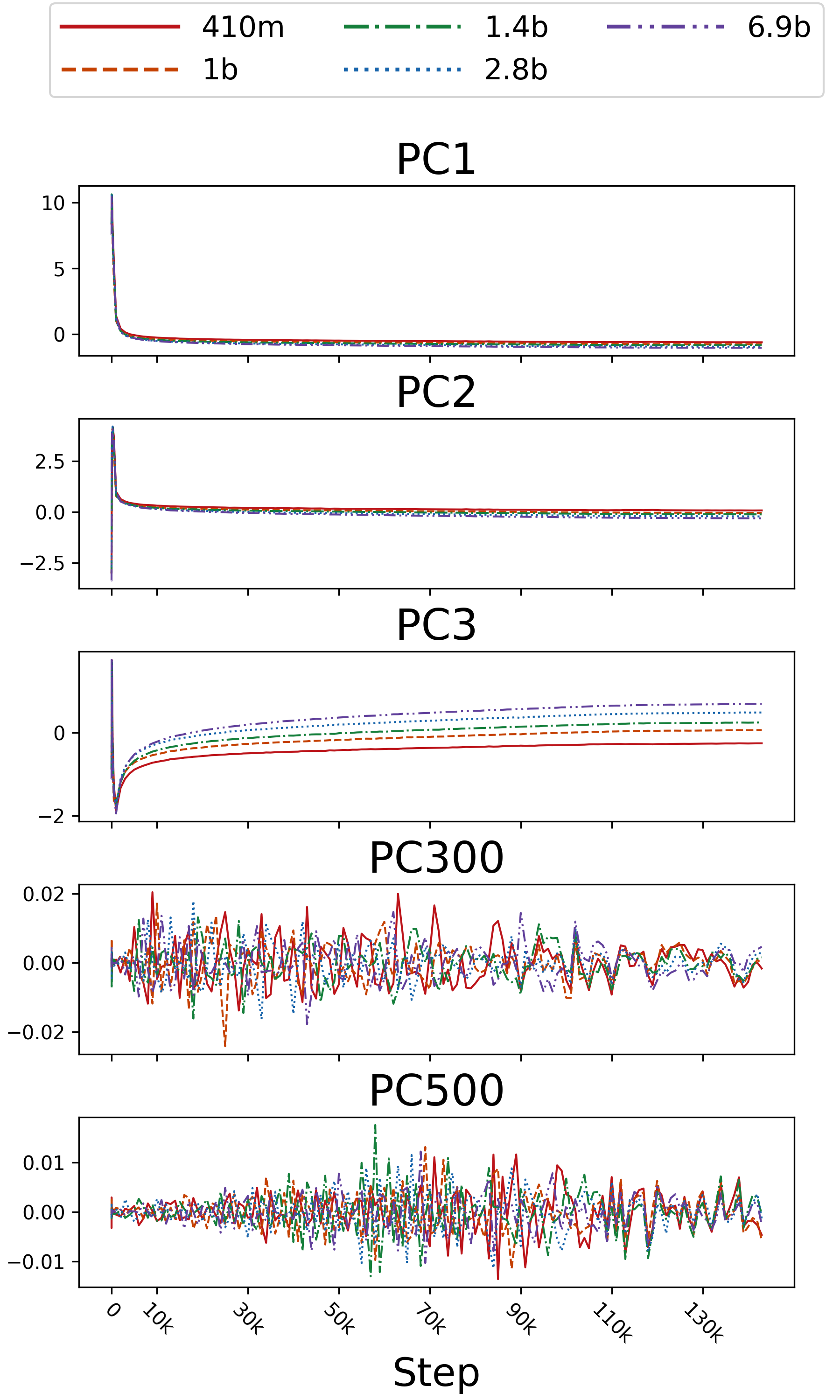}
    \caption{
    Variation of principal component values from PCA applied to the trajectories of all Pythia model sizes.
    }
    \label{fig:pretraining_pca_timeseries}
\end{figure}

We further analyze the same pretraining trajectories using principal component analysis (PCA) to highlight their global geometric structure.
In contrast to t-SNE, which emphasizes local neighborhood relationships, PCA captures large-scale variations shared across different model sizes.

Fig.~\ref{fig:pretraining_pca} shows the result of applying PCA simultaneously to all model sizes and plotting their trajectories. Fig.~\ref{fig:pretraining_pca_timeseries} further illustrates the evolution of multiple principal component values along the training steps.
Unlike t-SNE, the top principal components exhibit similar values across model sizes, resulting in overlapping trajectories. This difference likely arises from the fact that the top principal components capture global variation in the trajectories, whereas t-SNE emphasizes local structure.
In contrast, the lower principal components exhibit more noise-like behavior. As shown in Table~\ref{tab:pretraining_kl_size_pairwise}, the KL divergence between different model sizes at the final checkpoint is larger than that between consecutive saved checkpoints. This is consistent with the observation that trajectories are separated in the lower components.

\subsection{Limitations of Trajectory Visualization in Weight Space}
\label{app:weight_space_visualization}

\begin{figure}[t]
    \centering
    \includegraphics[width=1.0\linewidth]{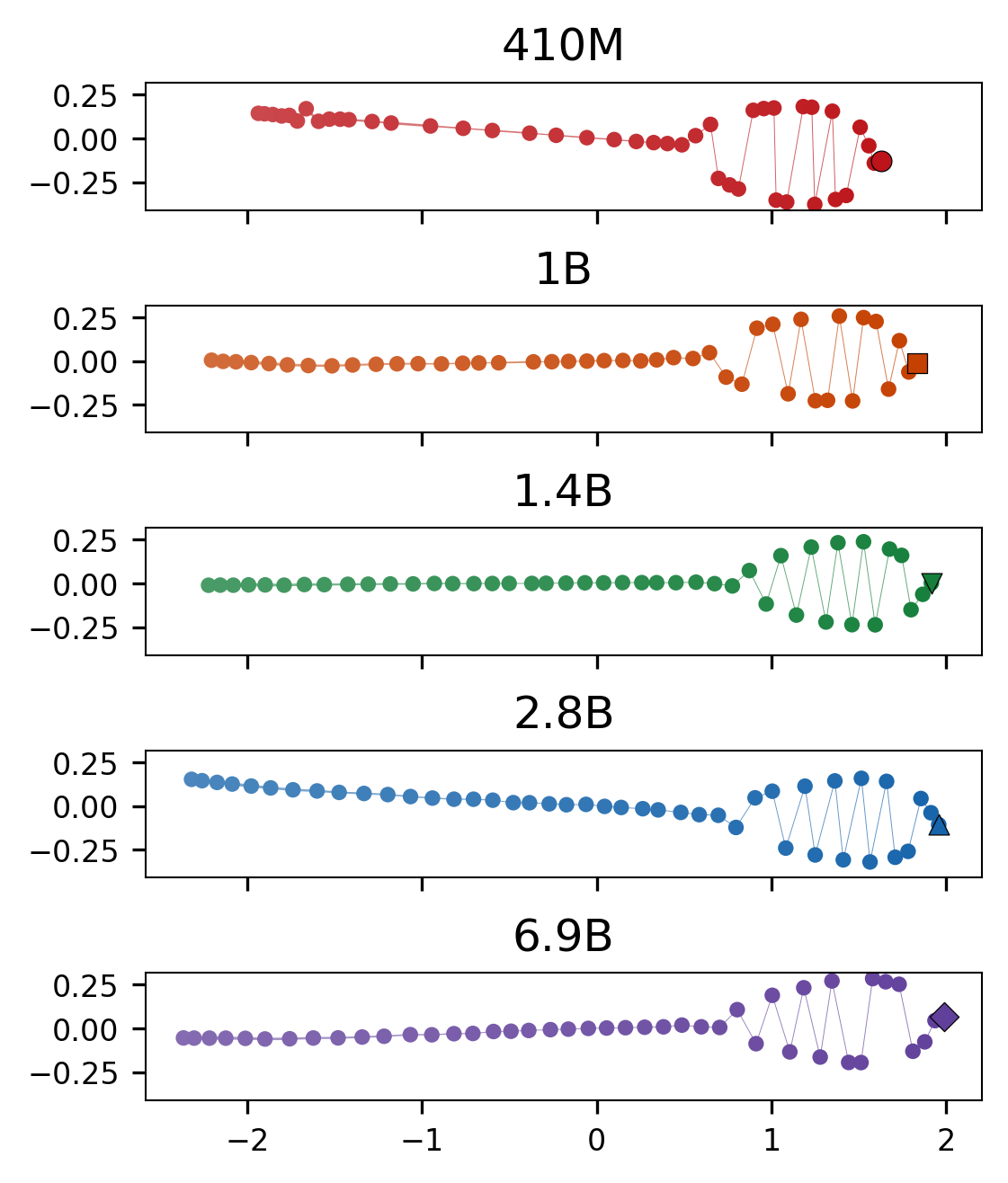}
    \caption{
    t-SNE visualization (perplexity=30) of the trajectories based on pairwise distances in the weight space for Pythia models after step 100k. To align plot ranges across model sizes, the trajectories were centered after t-SNE. They were also rotated to roughly align the directions of progression. Final checkpoints are indicated using distinct markers.
    }
    \label{fig:pretraining_weight_tsne}
\end{figure}

Since pretraining trajectories in the weight space do not admit a joint visualization across different model sizes and random seeds, we visualize them separately for each model.
For each model size, we computed a distance matrix containing the pairwise Euclidean distances between the weight vectors of all checkpoints. Figure~\ref{fig:pretraining_weight_tsne} shows the trajectories from step 100k onward, visualized using t-SNE\footnote{t-SNE can take a distance matrix as input.}. The trajectories were centered after t-SNE to align the plot ranges. In addition, for the 1B model, the trajectory was rotated by 180 degrees to roughly align its direction of progression with those of the other model sizes.

Despite these adjustments, the resulting visualizations remain difficult to interpret. Because each model requires a separate embedding, the relative positions and orientations of trajectories across models are arbitrary and cannot be meaningfully compared. This highlights a fundamental limitation of weight-space representations for trajectory visualization, in contrast to the log-likelihood space, which provides a shared coordinate system for joint analysis.

\section{Details of Sections~\ref{subsec:kl_training} and \ref{sec:pretraining}} \label{app:detail_pretraining}

To analyze the trajectories of language models during pretraining, we used Pythia models\footnote{Released under the Apache 2.0 License.} of various sizes, as listed in Table~\ref{tab:model_list_pretraining} in Appendix~\ref{app:model_list}. For each checkpoint, we computed log-likelihood vectors using the same set of 10,000 texts as in \citet{DBLP:conf/acl/OyamaYTS25}. The computations were performed on an NVIDIA RTX 6000 Ada and took approximately 10 minutes for a 7B model loaded in float16 precision.

\subsection{Preprocessing} \label{app:pretraining_prerprocess}

\begin{figure}[t]
    \centering
    \includegraphics[width=1.0\linewidth]{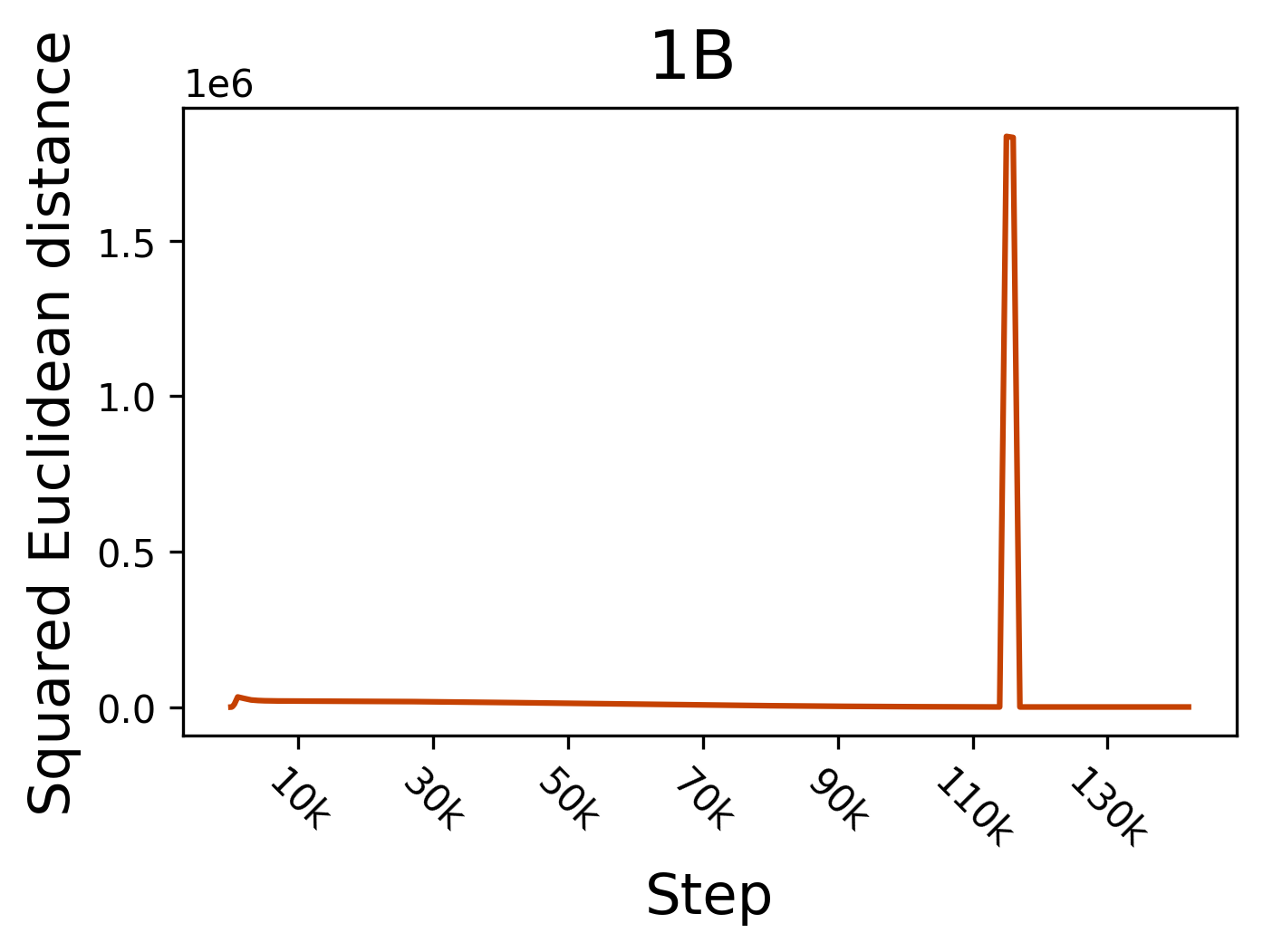}
    \caption{
    Squared Euclidean distance between the weights of consecutive saved checkpoints for Pythia 1B. The distances between 115k and 116k, and between 116k and 117k are abnormally large.
    }
    \label{fig:pretraining_weight_distance_1b}
\end{figure}

\begin{figure}[t]
    \centering
    \includegraphics[width=1.0\linewidth]{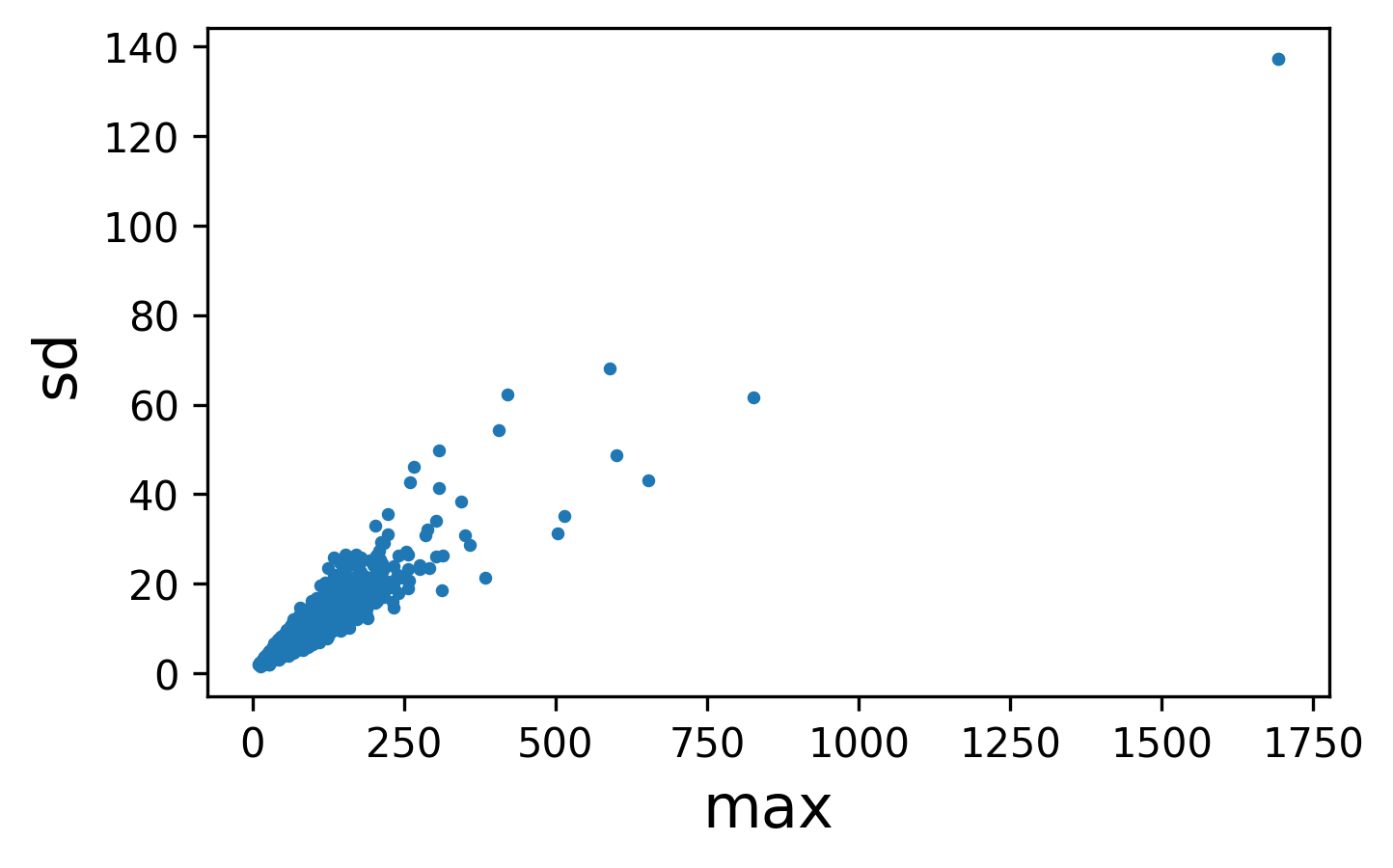}
    \caption{
    The relationship between the maximum and the standard deviation of the absolute differences in log-likelihoods between consecutive saved checkpoints of the same model size, computed for each text.
    }
    \label{fig:pretraining_outlier_text_score}
\end{figure}

\begin{figure}[!ht] 
    \centering
    \includegraphics[width=0.9\linewidth]{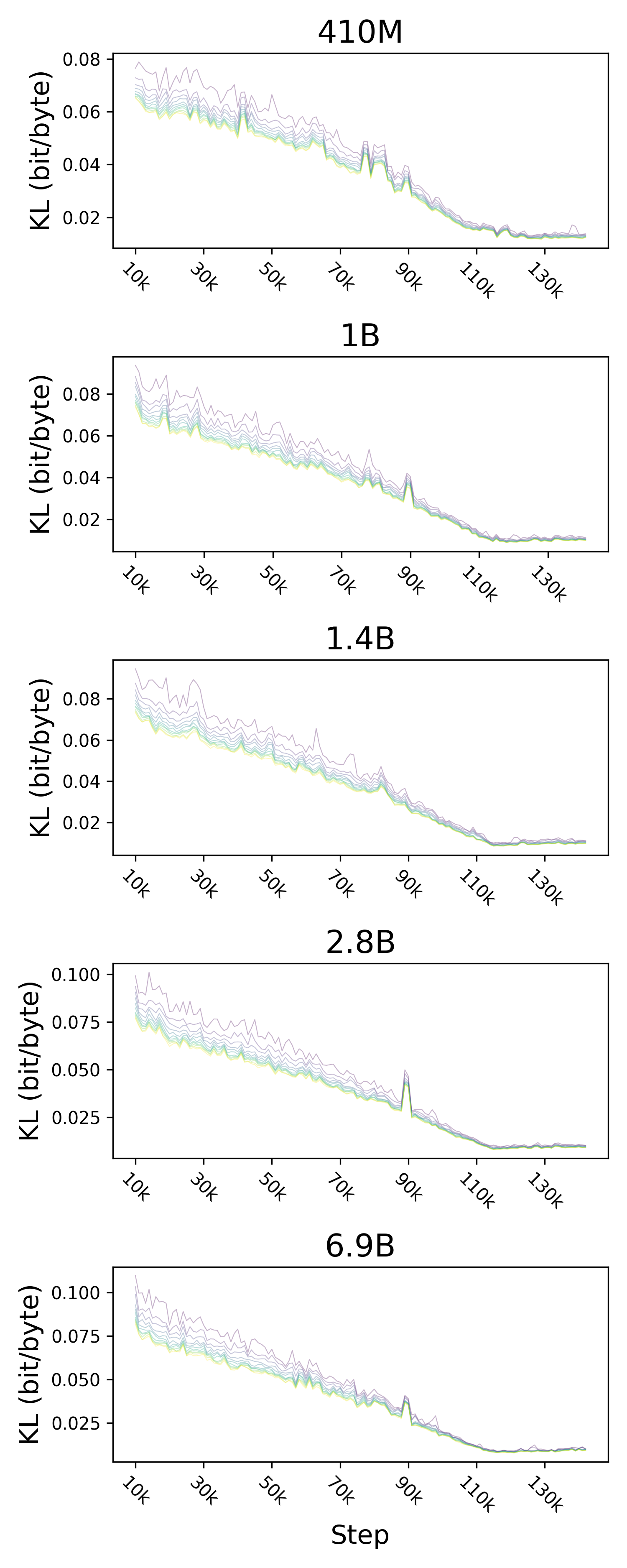}
    \caption{
    Change in KL divergence after step 10k for different numbers of top outlier texts removed (10, 100, 200, ..., 1000). Brighter colors indicate larger numbers of removed texts.
    }
    \label{fig:pretraining_outlier_kl}
\end{figure}

Assuming that the KL divergence between consecutive saved checkpoints changes continuously, we regarded certain checkpoints and texts as outliers. All experiments were conducted after removing these outliers. This appendix provides a detailed description of the preprocessing procedure.

\paragraph{Removing outlier models.}

As shown in Fig.~\ref{fig:pretraining_weight_distance_1b}, the Euclidean distance between weights near step 116k of Pythia 1B is abnormally large. We suspect that this checkpoint may not have been saved properly and therefore excluded it from our analysis.
A question about this checkpoint was also raised on the Discussions page of the Pythia 1B model on Hugging Face\footnote{\url{https://huggingface.co/EleutherAI/pythia-1b/discussions/4}}.

\paragraph{Removing outlier texts.}
Following \citet{DBLP:conf/acl/OyamaYTS25}, we clip the log-likelihood matrix $\bm{L}$ at the bottom 2\%, after excluding the outlier model.
Next, we consider the log-likelihood difference for each text between consecutive saved checkpoints after the warmup phase. For each text, we define its outlier score as the maximum absolute difference across all model sizes and training steps. Specifically, letting $p_{i,t}$ denote the model of size $i$ at step $t$, the outlier score for a text $x$ is given by $\max_{i, t} \left\{ \left| \log p_{i, t+1}(x) - \log p_{i, t}(x) \right| \right\}$.
It is also possible to assign scores to texts using the standard deviation of these differences. However, Fig.~\ref{fig:pretraining_outlier_text_score} shows that the maximum and the standard deviation are highly correlated. Therefore, we adopt the maximum value for scoring. The figure further confirms the presence of texts that clearly behave as outliers. For example, the text with the highest outlier score was a meaningless and repetitive sequence of symbols, such as \texttt{028a28a0028a28a0028a28a...}.

Fig.~\ref{fig:pretraining_outlier_kl} shows the KL divergence between consecutive saved checkpoints after step 10k\footnote{The maximum is computed from step 2k onward, after the warmup phase, but we visualize from step 10k to better capture the variation in KL divergence.}, shown for different numbers of top outlier-scored texts removed (top 10, 100, 200, ..., 1000).
When the top 300 texts (3\%) are removed, the variation in KL divergence stabilizes with respect to the number of removed texts. Therefore, we excluded these 300 texts from our experiments.

\subsection{KL Divergence between Checkpoints} \label{app:pretraining_kl}

\begin{figure}[!ht]
    \centering
    \includegraphics[width=0.9\linewidth]{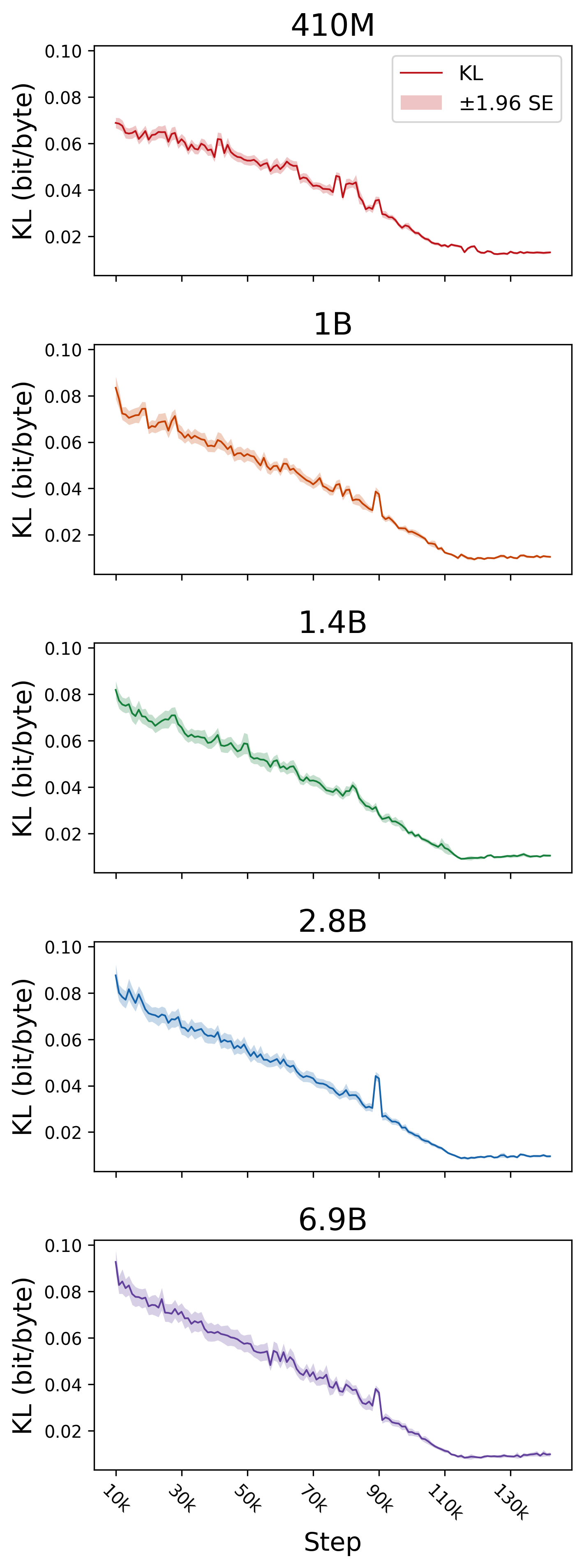}
    \caption{
    KL divergence and its standard error between consecutive saved checkpoints of Pythia after step 10k.
    }
    \label{fig:pretraining_kl_diff_se}
\end{figure}

\begin{table}[t]
\scriptsize
\centering
\begin{tabular}{lccccc}
\toprule
        & 410M & 1B   & 1.4B & 2.8B & 6.9B \\
\midrule
410M    & 0.00 & 0.28 & 0.52 & 1.04 & 1.7 \\
1B      & 0.28 & 0.00 & 0.15 & 0.44 & 0.92 \\
1.4B    & 0.52 & 0.15 & 0.00 & 0.21 & 0.57 \\
2.8B    & 1.0 & 0.44 & 0.21 & 0.00 & 0.21 \\
6.9B    & 1.7 & 0.92 & 0.57 & 0.21 & 0.00 \\
\bottomrule
\end{tabular}
\caption{
KL divergence (bits/byte) between different model sizes at the final checkpoint of Pythia.
}
\label{tab:pretraining_kl_size_pairwise}
\end{table}

During pretraining, the learning rate changes dynamically with the training step. In particular, the first 1,430 steps correspond to the warmup phase, during which the learning rate increases linearly. As a result, as shown in the left panel of Fig.~\ref{fig:trajectory_subsection}, the KL divergence between the checkpoints takes on relatively large values, mostly exceeding 10 bits/byte during this interval.
The right panel of Fig.~\ref{fig:trajectory_subsection} shows the variation in KL divergence between consecutive saved checkpoints for all model sizes. In this figure, the value at step 10k, for example, represents the KL divergence between the checkpoints at steps 10k and 11k.
Additionally, Fig.~\ref{fig:pretraining_kl_diff_se} plots the KL divergence between consecutive saved checkpoints after step 10k for Pythia, along with the corresponding standard error. The derivation of the standard error is described in Appendix~\ref{app:KL_SE}.
Finally, Table~\ref{tab:pretraining_kl_size_pairwise} presents the pairwise KL divergence between model sizes at the final checkpoint. As discussed in Section~\ref{sec:pretraining}, the divergence caused by differences in model size is larger than that resulting from 1k steps of training.

\subsection{Estimation of Diffusion Exponents}
\label{app:diffusion_estimation}

\begin{table}[t]
\scriptsize
\centering
\begin{tabular}{lrr}
\toprule
Size                  & Log-likelihood & Weight   \\
\midrule
\multirow{2}{*}{410M} & 0.849          & 0.993    \\
                      & ($\pm $0.158)          & ($\pm $0.00622) \\
\multirow{2}{*}{1B}   & 0.935          & 0.999    \\
                      & ($\pm$ 0.0866)        & ($\pm $0.000922)\\
\multirow{2}{*}{1.4B} & 0.955          & 0.999    \\
                      & ($\pm$ 0.0336)         & ($\pm $0.000687)\\
\multirow{2}{*}{2.8B} & 0.959          & 0.999    \\
                      & ($\pm$ 0.088)          & ($\pm $0.000704)\\
\multirow{2}{*}{6.9B} & 0.977          & 0.999    \\
                      & ($\pm$ 0.0305)         & ($\pm $0.000975)\\
\bottomrule
\end{tabular}
\caption{
The mean and standard deviation of the coefficient of determination ($R^2$) when varying the starting step used to compute the diffusion exponent.
}
\label{tab:pretraining_diffusion_exponent_R}
\end{table}

In this subsection, we describe the procedure for estimating diffusion exponents in both the log-likelihood and weight spaces.
The left panel of Fig.~\ref{fig:diffusion_subsection} shows the evolution of squared Euclidean distances in weights and log-likelihood vectors over 10k steps starting from step 120k. The right panel of Fig.~\ref{fig:diffusion_subsection} illustrates how the estimated diffusion exponent changes when varying the starting step for the calculation. To ensure stability, we vary the starting point only after the warmup phase.
The exponent is estimated using linear regression (least squares) over a 10k-step window beginning at each initial step. Summary statistics of the coefficient of determination ($R^2$) for each regression are provided in Table~\ref{tab:pretraining_diffusion_exponent_R}.

\subsection{Different Seeds} \label{app:pretraining_seed}

\begin{figure}[t]
    \centering
    \includegraphics[width=0.9\linewidth]{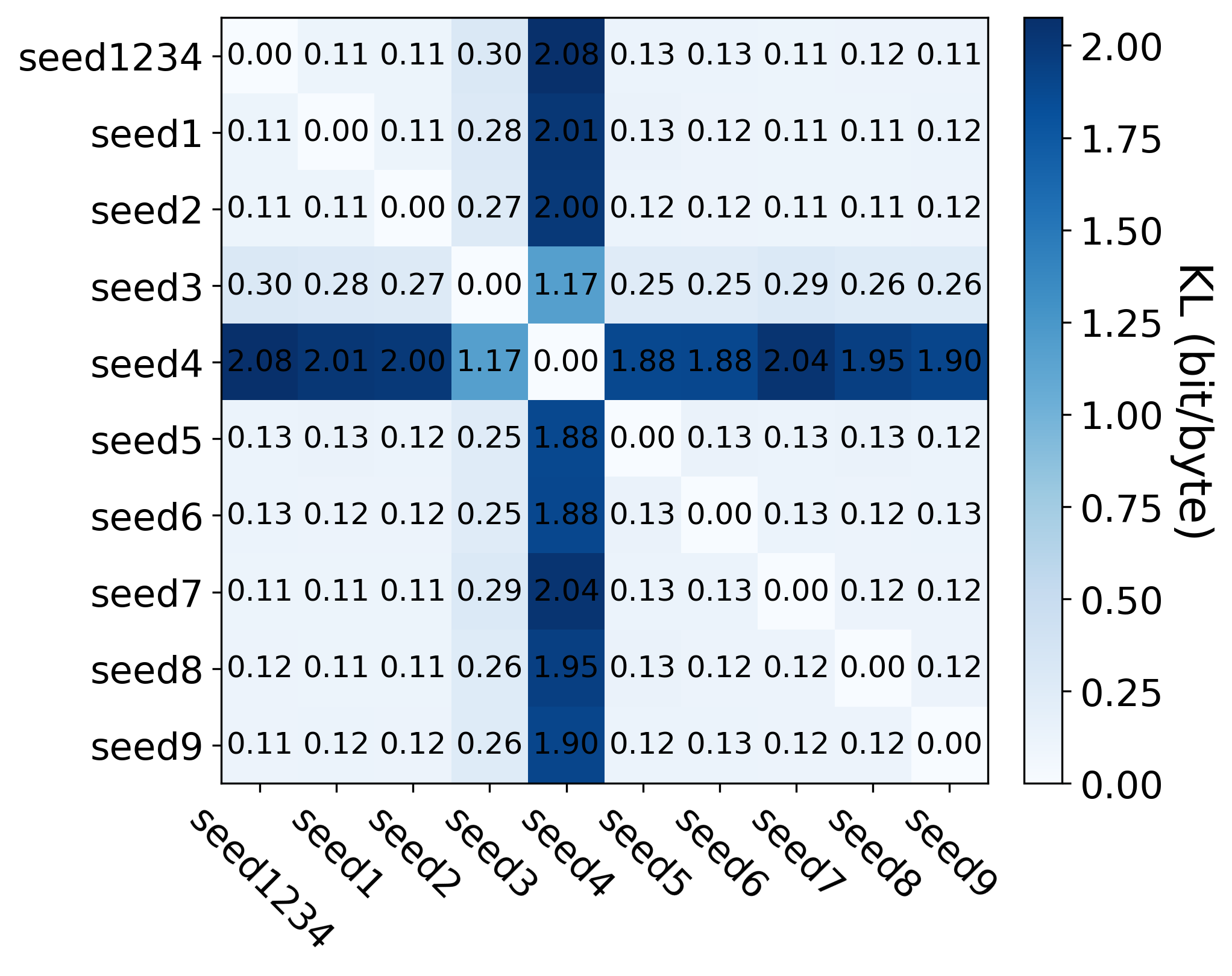}
    \caption{
    KL divergence between the final checkpoints of Pythia 410M models with different seeds.
    }
    \label{fig:pretraining_seed_kl_last}
\end{figure}

\citet{wal2025polypythias} trained Pythia models with nine different seeds and released checkpoints at the same steps as in the original Pythia suite.
The random seed affects the initialization of weights and the order of the training data.
We used all available checkpoints for the 410M models with seeds 1 through 9. However, at the time of writing this paper, the weights for (seed, step) = (2, 111k), (6, 88k), and (9, 40k) were not available on Hugging Face and were therefore excluded from our analysis\footnote{For example, see \url{https://huggingface.co/EleutherAI/pythia-410m-seed2/tree/step111000}.}. The original 410M model was trained with seed 1234.

Fig.~\ref{fig:pretraining_seed_kl_last} shows the KL divergence between the final checkpoints of models trained with different seeds. Despite sharing the same architecture and data, the KL divergence between seed 3 and the other seeds, as well as between seed 4 and the others, are significantly larger.
According to \citet{wal2025polypythias}, models trained with seed 3 and seed 4 experienced loss spikes during training and are considered outliers in terms of performance. This suggests that KL divergence is also effective in identifying such outlier models.
Therefore, seed 3 and seed 4 are excluded when generating Fig.~\ref{fig:pretraining_tsne_std}. We also report in Table~\ref{tab:pretraining_kl} the average KL divergence between the final checkpoint of the original 410M model and those of seeds 1, 2, 5, 6, 7, 8, and 9.

\section{Details of Section~\ref{subsec:quantization}}\label{app:quantization}
Here, we describe the language models used in Section~\ref{subsec:quantization}. We selected the 50 most downloaded models from those analyzed in \citet{DBLP:conf/acl/OyamaYTS25}. A complete list of the models is provided in Table~\ref{tab:model_list_layer} in Appendix~\ref{app:model_list}.

For the quantization methods, we applied 8-bit~\cite{dettmers2022llmint8} and 4-bit~\cite{arxiv:2305.14314} quantization using \texttt{bitsandbytes}\footnote{\url{https://github.com/bitsandbytes-foundation/bitsandbytes}}.

\section{Details of Section~\ref{subsec:fine-tuning}} \label{app:detail_finetuning}

We describe the language models used in Section~\ref{subsec:fine-tuning}. From the 1,018 models analyzed in \citet{DBLP:conf/acl/OyamaYTS25}, we identified fine-tuned models as those whose base model is specified in the metadata available via the Hugging Face Hub API\footnote{\url{https://github.com/huggingface/huggingface_hub}}. Our analysis was conducted on pairs of fine-tuned models and their base models. Models associated with multiple base models (i.e., merged models) were excluded. A list of the language models used in this analysis is provided in Table~\ref{tab:model_list_ft} in Appendix~\ref{app:model_list}. In Fig.~\ref{fig:ft_layer_tsne}b, there are a few pairs connected by line segments that span across different colors, namely different model types. This occurs because, under the model type classification method used in \citet{DBLP:conf/acl/OyamaYTS25}, some fine-tuned models were categorized as belonging to the ``others'' type.

Summary statistics for these distributions are reported in Table~\ref{tab:kl_scale}. Fig.~\ref{fig:kl_cheat_sheet} shows that KL divergence after fine-tuning tends to be slightly smaller, as indicated by the left-skewed distribution.
As noted in \citet{DBLP:conf/acl/OyamaYTS25}, models of the same type tend to have smaller KL divergence. To more accurately assess the effect of fine-tuning, we also analyze random model pairs sampled within the same model type.

\section{Details of Section~\ref{subsec:layer}} \label{app:detail_layer}

\begin{figure}[t]
    \centering
    \includegraphics[width=1.0\linewidth]{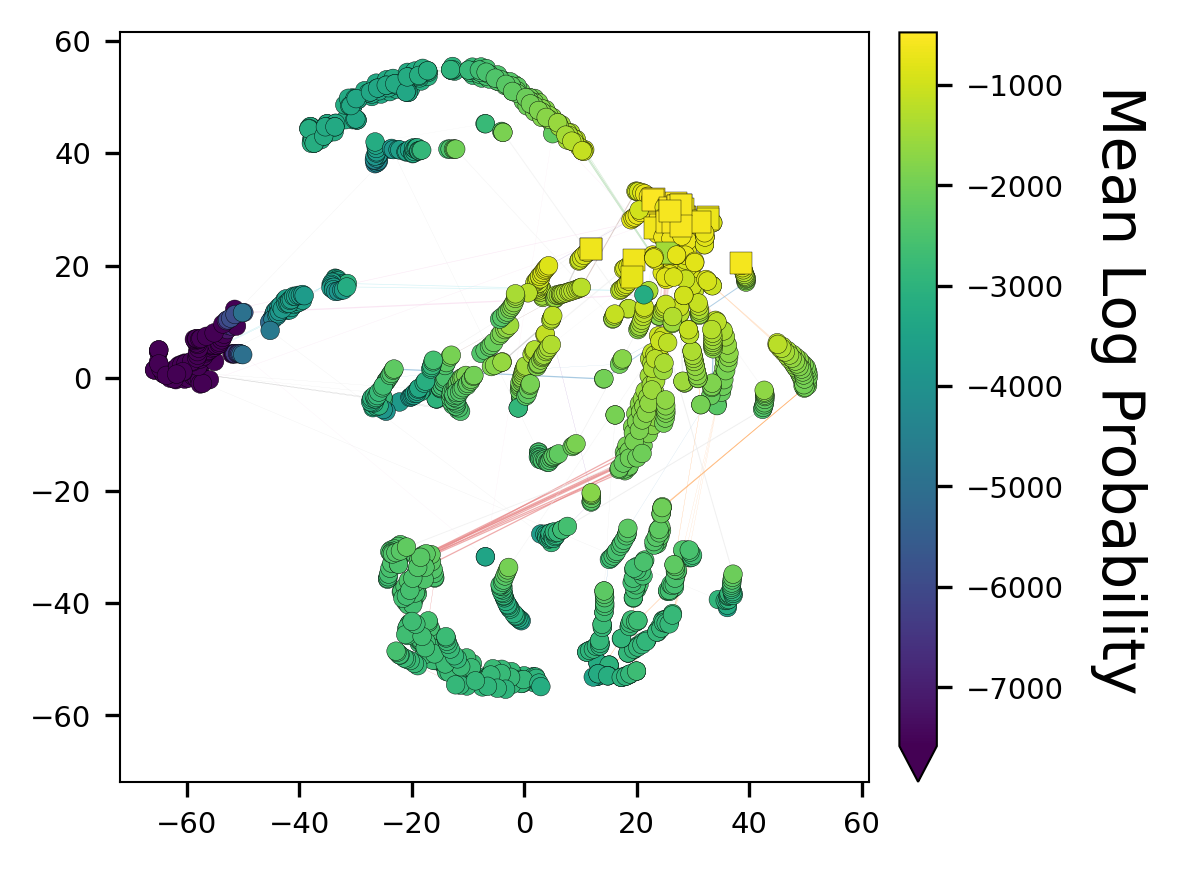}
    \caption{
    Re-rendering of the top panel of Fig.~\ref{fig:ft_layer_tsne}c. Colors indicate mean log-likelihood, clipped at the bottom 5\% across all values to improve visibility.
    }
    \label{fig:layer_meanlogp}
\end{figure}

\begin{figure}[t]
    \centering
    \includegraphics[width=0.9\linewidth]{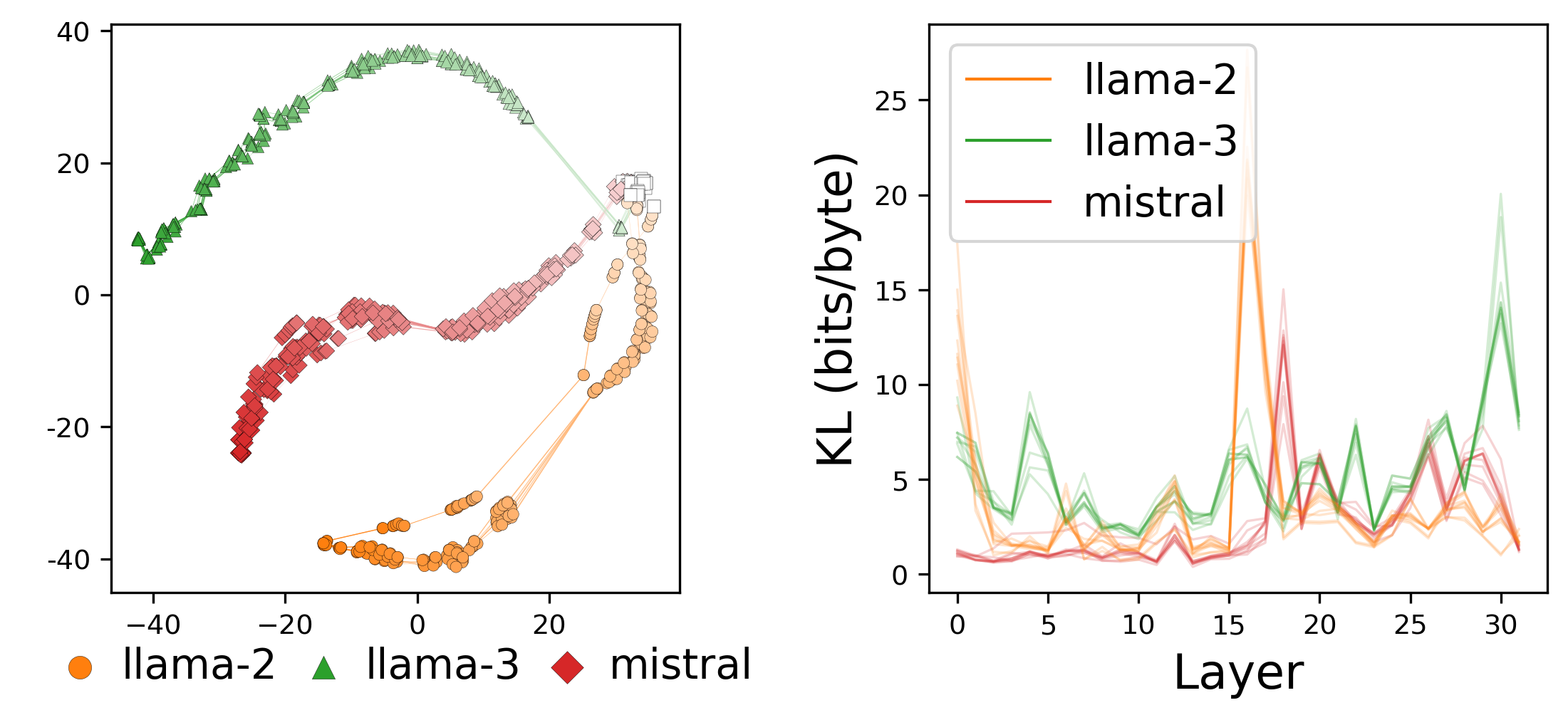}
    \caption{
    For 10 models per model type: (left) t-SNE visualization of trajectories across layers, where the final layer is indicated by a white square; (right) KL divergence between adjacent layers (corresponding to Table~\ref{tab:layer_kl}). 
    }
    \label{fig:layer_kl}
    \vspace{-0.3cm}
\end{figure}

\begin{figure}[t]
    \centering
    \includegraphics[width=0.9\linewidth]{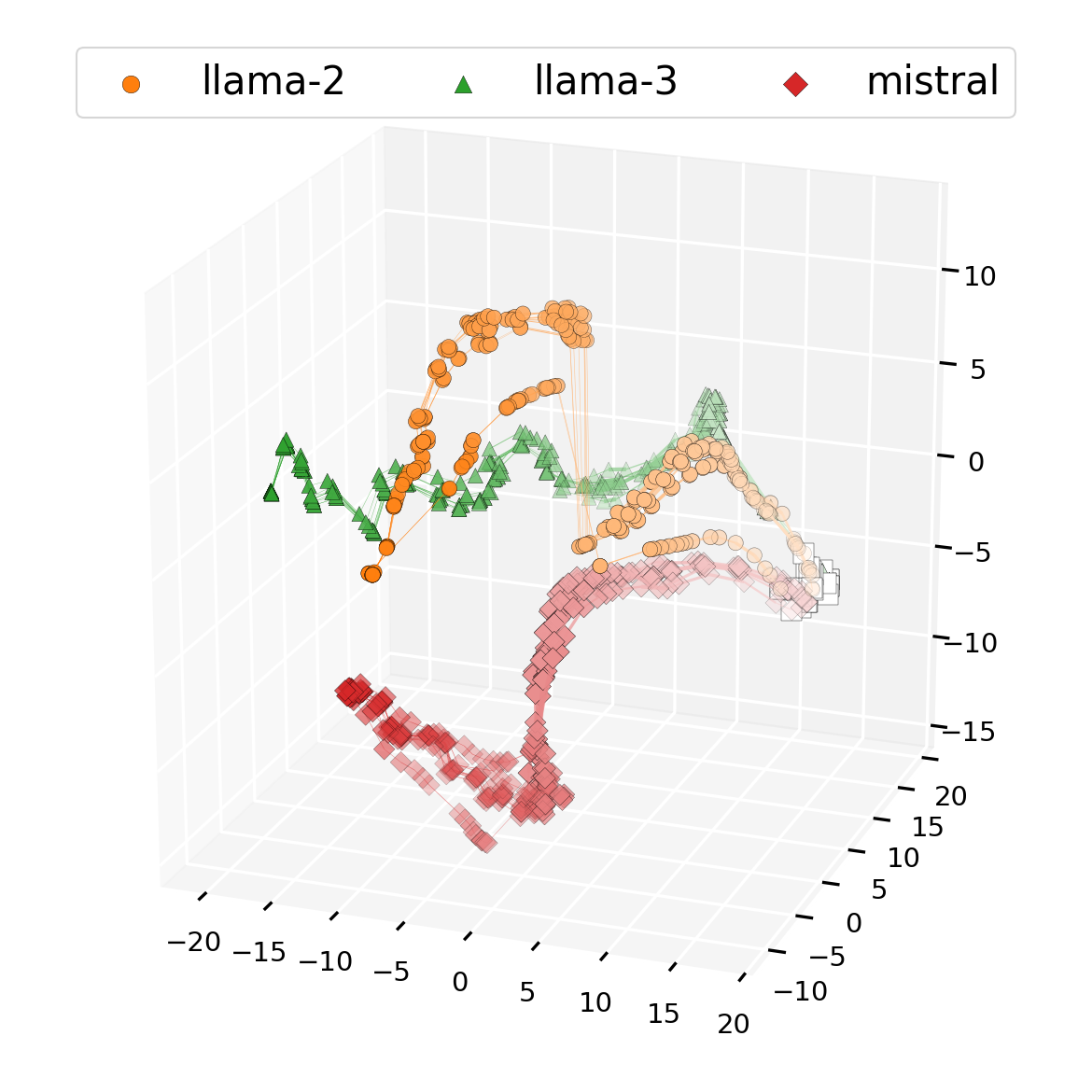}
    \caption{
    3D t-SNE visualization (perplexity=30) of the left panel of Fig.~\ref{fig:layer_kl}. Deeper layers are represented with lighter colors, and the final layer is indicated by a white square.
    }
    \vspace{-0.3cm}
    \label{fig:ft_layer_tsne_3d}
\end{figure}

Here, we describe the language models used in Section~\ref{subsec:layer}. For the trajectories shown in Fig.~\ref{fig:ft_layer_tsne}c and Fig.~\ref{fig:layer_meanlogp}, we selected the 50 most downloaded models from those analyzed in \citet{DBLP:conf/acl/OyamaYTS25}.
A complete list of the models is provided in Table~\ref{tab:model_list_layer} in Appendix~\ref{app:model_list}.

Fig.~\ref{fig:layer_meanlogp} shows the trajectories of models across layers, color-coded by mean log-likelihood, depending on layer depth. The mean log-likelihood increases as the layer depth increases. This indicates that the direction of trajectory progression corresponds to deeper layers, that is, to increasing mean log-likelihood.

Additionally, for detailed analysis, we selected models with 32 layers from the llama-2, llama-3, and mistral types. For each of these types, we used the 10 most downloaded models. That is, the top 50 do not include 10 models for each of llama-2, llama-3, and mistral types. Therefore, additional models were included to ensure 10 models per category, resulting in a total of 9 added models. Fig.~\ref{fig:layer_kl} shows the trajectories of these models and the KL divergence between adjacent layers, while Fig.~\ref{fig:ft_layer_tsne_3d} illustrates the trajectories in 3d space. For all model types, the trajectories are already clustered by model type from the shallow layers onward. In addition, the KL divergence values exhibit similar trends within each type.

\section{Model List}\label{app:model_list}

The models used in Sections~\ref{subsec:kl_training}, \ref{subsec:fine-tuning}, and~\ref{subsec:layer} are listed in Tables~\ref{tab:model_list_pretraining}, \ref{tab:model_list_ft}, and~\ref{tab:model_list_layer}, respectively.  
Table~\ref{tab:model_list_pretraining} is sorted by model size; Table~\ref{tab:model_list_ft} is sorted alphabetically by model name, considering the model generation, model type, and corresponding base model; and Table~\ref{tab:model_list_layer} is sorted in descending order of download counts.

BibTeX entries for each model were selected according to the procedure described by \citet{DBLP:conf/acl/OyamaYTS25}.  
Note that, for the Pythia models trained with different random seeds~\cite{wal2025polypythias} and used in Section~\ref{sec:pretraining}, the corresponding BibTeX entries in Table~\ref{tab:model_list_pretraining} were prepared manually.

\onecolumn

{\scriptsize
\begin{longtable}{r@{\hspace{1em}}l}
\caption{List of 14 models in the experiments for pretraining models. ``ID'' denotes the index sorted by model size; ``Model Name'' denotes the name of the model. Seeds 3 and 4 of Pythia-410M are excluded from the experiments in Sections~\ref{subsec:kl_training} and \ref{sec:pretraining}.}\\

\toprule
ID & Model Name\\
\midrule
\endfirsthead
    
\toprule
ID & Model Name\\
\midrule
\endhead
1 & \parbox[t]{8cm}{EleutherAI/pythia-410m~\cite{arxiv:2101.00027,arxiv:2201.07311,arxiv:2304.01373}} \\
2 & \parbox[t]{8cm}{EleutherAI/pythia-410m-seed1~\cite{wal2025polypythias}} \\
3 & \parbox[t]{8cm}{EleutherAI/pythia-410m-seed2~\cite{wal2025polypythias}} \\
4 & \parbox[t]{8cm}{EleutherAI/pythia-410m-seed3~\cite{wal2025polypythias}} \\
5 & \parbox[t]{8cm}{EleutherAI/pythia-410m-seed4~\cite{wal2025polypythias}} \\
6 & \parbox[t]{8cm}{EleutherAI/pythia-410m-seed5~\cite{wal2025polypythias}} \\
7 & \parbox[t]{8cm}{EleutherAI/pythia-410m-seed6~\cite{wal2025polypythias}} \\
8 & \parbox[t]{8cm}{EleutherAI/pythia-410m-seed7~\cite{wal2025polypythias}} \\
9 & \parbox[t]{8cm}{EleutherAI/pythia-410m-seed8~\cite{wal2025polypythias}} \\
10 & \parbox[t]{8cm}{EleutherAI/pythia-410m-seed9~\cite{wal2025polypythias}} \\
11 & \parbox[t]{8cm}{EleutherAI/pythia-1b~\cite{arxiv:2101.00027,arxiv:2201.07311,arxiv:2304.01373}} \\
12 & \parbox[t]{8cm}{EleutherAI/pythia-1.4b~\cite{arxiv:2101.00027,arxiv:2201.07311,arxiv:2304.01373}} \\
13 & \parbox[t]{8cm}{EleutherAI/pythia-2.8b~\cite{arxiv:2101.00027,arxiv:2201.07311,arxiv:2304.01373}} \\
14 & \parbox[t]{8cm}{EleutherAI/pythia-6.9b~\cite{arxiv:2101.00027,arxiv:2201.07311,arxiv:2304.01373}} \\

\bottomrule
\label{tab:model_list_pretraining}
\end{longtable}
}

{\scriptsize
\begin{longtable}{r@{\hspace{1em}}l@{\hspace{1em}}l@{\hspace{1em}}l@{\hspace{1em}}r@{\hspace{1em}}r}
\caption{List of 87 models in the experiments for fine-tuning models. ``ID'' denotes the alphabetical index considering the generation and the base model; ``Model Name'' denotes the name of the model; ``Base Model Name'' denotes the name of the base model before training; ``Model Type'' denotes the model classification defined in \citet{DBLP:conf/acl/OyamaYTS25}; ``Generation'' denotes the model generation in the training process; ``DLs'' denotes the total number of downloads.}\\

\toprule
ID & Model Name & Base Model Name & Model Type & Generation & DLs\\
\midrule
\endfirsthead
    
\toprule
ID & Model Name & Base Model Name & Model Type & Generation & DLs\\
\midrule
\endhead
1 & \parbox[t]{6cm}{deepseek-ai/DeepSeek-Prover-V1.5-Base~\cite{arxiv:2408.08152}} & -- & deepseek & 1 & 230 \\
2 & \parbox[t]{6cm}{tiiuae/falcon-rw-1b~\cite{arxiv:2005.14165,arxiv:2108.12409,arxiv:2205.14135,arxiv:2306.01116}} & -- & falcon & 1 & 22921 \\
3 & \parbox[t]{6cm}{google/gemma-1.1-7b-it~\cite{arxiv:1705.03551,arxiv:1809.02789,arxiv:1804.06876,arxiv:1905.10044,arxiv:1811.00937,arxiv:1904.09728,arxiv:1907.10641,arxiv:1911.11641,arxiv:1911.01547,arxiv:1905.07830,arxiv:2009.03300,arxiv:2107.03374,arxiv:2108.07732,arxiv:2110.14168,arxiv:2110.08193,arxiv:2206.04615,arxiv:2304.06364,arxiv:2312.11805}} & -- & gemma & 1 & 19835 \\
4 & \parbox[t]{6cm}{google/gemma-7b~\cite{arxiv:1705.03551,arxiv:1809.02789,arxiv:1804.06876,arxiv:1804.09301,arxiv:1911.11641,arxiv:1911.01547,arxiv:1905.10044,arxiv:1811.00937,arxiv:1904.09728,arxiv:1905.07830,arxiv:1907.10641,arxiv:2009.11462,arxiv:2107.03374,arxiv:2108.07732,arxiv:2009.03300,arxiv:2101.11718,arxiv:2110.14168,arxiv:2109.07958,arxiv:2203.09509,arxiv:2110.08193,arxiv:2206.04615,arxiv:2304.06364,arxiv:2305.14314,arxiv:2312.11805}} & -- & gemma & 1 & 72047 \\
5 & \parbox[t]{6cm}{openchat/openchat-3.5-0106-gemma~\cite{arxiv:2309.11235}} & -- & gemma & 1 & 7817 \\
6 & \parbox[t]{6cm}{openlm-research/open\_llama\_3b~\cite{together2023redpajama,arxiv:2302.13971,openlm2023openllama}} & -- & llama-1 & 1 & 157405 \\
7 & \parbox[t]{6cm}{codellama/CodeLlama-13b-Instruct-hf~\cite{arxiv:2308.12950}} & -- & llama-2 & 1 & 25248 \\
8 & \parbox[t]{6cm}{meta-llama/Llama-2-13b-hf~\cite{arxiv:2307.09288}} & -- & llama-2 & 1 & 120871 \\
9 & \parbox[t]{6cm}{meta-llama/Llama-2-7b-hf~\cite{arxiv:2307.09288}} & -- & llama-2 & 1 & 1294737 \\
10 & \parbox[t]{6cm}{meta-llama/Meta-Llama-3-8B-Instruct~\cite{llama3modelcard}} & -- & llama-3 & 1 & 2058011 \\
11 & \parbox[t]{6cm}{meta-llama/Meta-Llama-3-8B~\cite{llama3modelcard}} & -- & llama-3 & 1 & 675850 \\
12 & \parbox[t]{6cm}{NousResearch/Hermes-2-Pro-Llama-3-8B~\cite{Hermes-2-Pro-Llama-3-8B}} & -- & llama-3 & 1 & 26797 \\
13 & \parbox[t]{6cm}{mistralai/Mistral-7B-Instruct-v0.2~\cite{arxiv:2310.06825}} & -- & mistral & 1 & 3565248 \\
14 & \parbox[t]{6cm}{mistralai/Mistral-7B-v0.1~\cite{arxiv:2310.06825}} & -- & mistral & 1 & 1750089 \\
15 & \parbox[t]{6cm}{mlabonne/Marcoro14-7B-slerp} & -- & mistral & 1 & 3792 \\
16 & \parbox[t]{6cm}{mlabonne/NeuralMonarch-7B} & -- & mistral & 1 & 13486 \\
17 & \parbox[t]{6cm}{stabilityai/japanese-stablelm-base-gamma-7b~\cite{arxiv:2310.06825}} & -- & mistral & 1 & 2072 \\
18 & \parbox[t]{6cm}{beomi/KoRWKV-6B} & -- & others & 1 & 2127 \\
19 & \parbox[t]{6cm}{microsoft/Orca-2-13b~\cite{arxiv:2311.11045}} & -- & others & 1 & 13616 \\
20 & \parbox[t]{6cm}{upstage/SOLAR-10.7B-v1.0~\cite{arxiv:2312.15166}} & -- & others & 1 & 24478 \\
21 & \parbox[t]{6cm}{01-ai/Yi-1.5-9B~\cite{arxiv:2403.04652}} & -- & others & 1 & 20252 \\
22 & \parbox[t]{6cm}{deepseek-ai/DeepSeek-Prover-V1.5-SFT~\cite{arxiv:2408.08152}} & deepseek-ai/DeepSeek-Prover-V1.5-Base & deepseek & 2 & 6459 \\
23 & \parbox[t]{6cm}{euclaise/falcon\_1b\_stage1} & tiiuae/falcon-rw-1b & falcon & 2 & 2119 \\
24 & \parbox[t]{6cm}{lemon-mint/gemma-ko-7b-instruct-v0.71} & google/gemma-1.1-7b-it & gemma & 2 & 2232 \\
25 & \parbox[t]{6cm}{lemon-mint/gemma-7b-openhermes-v0.80} & google/gemma-1.1-7b-it & gemma & 2 & 4867 \\
26 & \parbox[t]{6cm}{google/gemma-7b-it~\cite{arxiv:1705.03551,arxiv:1809.02789,arxiv:1804.06876,arxiv:1804.09301,arxiv:1911.11641,arxiv:1911.01547,arxiv:1905.10044,arxiv:1811.00937,arxiv:1904.09728,arxiv:1905.07830,arxiv:1907.10641,arxiv:2009.11462,arxiv:2107.03374,arxiv:2108.07732,arxiv:2009.03300,arxiv:2101.11718,arxiv:2110.14168,arxiv:2109.07958,arxiv:2203.09509,arxiv:2110.08193,arxiv:2206.04615,arxiv:2304.06364,arxiv:2312.11805}} & google/gemma-7b & gemma & 2 & 66776 \\
27 & \parbox[t]{6cm}{Telugu-LLM-Labs/Indic-gemma-7b-finetuned-sft-Navarasa-2.0} & google/gemma-7b & gemma & 2 & 2025 \\
28 & \parbox[t]{6cm}{lemon-mint/gemma-ko-7b-instruct-v0.62} & openchat/openchat-3.5-0106-gemma & gemma & 2 & 7543 \\
29 & \parbox[t]{6cm}{rinna/youri-7b~\cite{gpt-neox-library,arxiv:2307.09288,rinna-youri-7b,arxiv:2404.01657}} & meta-llama/Llama-2-7b-hf & llama-2 & 2 & 2512 \\
30 & \parbox[t]{6cm}{aaditya/Llama3-OpenBioLLM-8B~\cite{arxiv:2212.13138,arxiv:2303.13375,arxiv:2305.09617,OpenBioLLMs,arxiv:2402.07023,arxiv:2305.18290}} & meta-llama/Meta-Llama-3-8B & llama-3 & 2 & 9308 \\
31 & \parbox[t]{6cm}{cognitivecomputations/dolphin-2.9-llama3-8b} & meta-llama/Meta-Llama-3-8B & llama-3 & 2 & 130977 \\
32 & \parbox[t]{6cm}{cognitivecomputations/dolphin-2.9.1-llama-3-8b} & meta-llama/Meta-Llama-3-8B & llama-3 & 2 & 7575 \\
33 & \parbox[t]{6cm}{DeepMount00/Llama-3-8b-Ita} & meta-llama/Meta-Llama-3-8B & llama-3 & 2 & 179401 \\
34 & \parbox[t]{6cm}{dfurman/Llama-3-8B-Orpo-v0.1} & meta-llama/Meta-Llama-3-8B & llama-3 & 2 & 5146 \\
35 & \parbox[t]{6cm}{johnsnowlabs/JSL-MedLlama-3-8B-v2.0} & meta-llama/Meta-Llama-3-8B & llama-3 & 2 & 11813 \\
36 & \parbox[t]{6cm}{jondurbin/bagel-8b-v1.0} & meta-llama/Meta-Llama-3-8B & llama-3 & 2 & 7960 \\
37 & \parbox[t]{6cm}{openchat/openchat-3.6-8b-20240522~\cite{arxiv:2309.11235}} & meta-llama/Meta-Llama-3-8B & llama-3 & 2 & 10464 \\
38 & \parbox[t]{6cm}{rinna/llama-3-youko-8b~\cite{gpt-neox-library,rinna-llama-3-youko-8b,llama3modelcard,arxiv:2404.01657}} & meta-llama/Meta-Llama-3-8B & llama-3 & 2 & 1468 \\
39 & \parbox[t]{6cm}{ruslanmv/Medical-Llama3-8B} & meta-llama/Meta-Llama-3-8B & llama-3 & 2 & 5457 \\
40 & \parbox[t]{6cm}{lightblue/suzume-llama-3-8B-multilingual~\cite{arxiv:2405.12612}} & meta-llama/Meta-Llama-3-8B-Instruct & llama-3 & 2 & 16442 \\
41 & \parbox[t]{6cm}{MaziyarPanahi/Llama-3-8B-Instruct-v0.4} & meta-llama/Meta-Llama-3-8B-Instruct & llama-3 & 2 & 1407 \\
42 & \parbox[t]{6cm}{PathFinderKR/Waktaverse-Llama-3-KO-8B-Instruct~\cite{waktaversellama3modelcard,llama3modelcard}} & meta-llama/Meta-Llama-3-8B-Instruct & llama-3 & 2 & 2305 \\
43 & \parbox[t]{6cm}{shenzhi-wang/Llama3-8B-Chinese-Chat~\cite{shenzhi-wang-2024}} & meta-llama/Meta-Llama-3-8B-Instruct & llama-3 & 2 & 55718 \\
44 & \parbox[t]{6cm}{SJ-Donald/llama3-passthrough-chat} & meta-llama/Meta-Llama-3-8B-Instruct & llama-3 & 2 & 2241 \\
45 & \parbox[t]{6cm}{swap-uniba/LLaMAntino-3-ANITA-8B-Inst-DPO-ITA~\cite{arxiv:2312.09993,llama3modelcard,arxiv:2405.07101}} & meta-llama/Meta-Llama-3-8B-Instruct & llama-3 & 2 & 5995 \\
46 & \parbox[t]{6cm}{NousResearch/Hermes-2-Theta-Llama-3-8B~\cite{Hermes-2-Theta-Llama-3-8B}} & NousResearch/Hermes-2-Pro-Llama-3-8B & llama-3 & 2 & 12392 \\
47 & \parbox[t]{6cm}{vicgalle/Configurable-Hermes-2-Pro-Llama-3-8B~\cite{arxiv:2404.00495}} & NousResearch/Hermes-2-Pro-Llama-3-8B & llama-3 & 2 & 10273 \\
48 & \parbox[t]{6cm}{abacusai/bigstral-12b-32k} & mistralai/Mistral-7B-Instruct-v0.2 & mistral & 2 & 5216 \\
49 & \parbox[t]{6cm}{Mihaiii/Metis-0.3} & mistralai/Mistral-7B-Instruct-v0.2 & mistral & 2 & 1279 \\
50 & \parbox[t]{6cm}{alignment-handbook/zephyr-7b-sft-full} & mistralai/Mistral-7B-v0.1 & mistral & 2 & 11605 \\
51 & \parbox[t]{6cm}{cognitivecomputations/dolphin-2.2.1-mistral-7b} & mistralai/Mistral-7B-v0.1 & mistral & 2 & 7292 \\
52 & \parbox[t]{6cm}{cypienai/cymist-2-v02-SFT~\cite{arxiv:1910.09700}} & mistralai/Mistral-7B-v0.1 & mistral & 2 & 2717 \\
53 & \parbox[t]{6cm}{HuggingFaceH4/mistral-7b-sft-beta} & mistralai/Mistral-7B-v0.1 & mistral & 2 & 7408 \\
54 & \parbox[t]{6cm}{HuggingFaceH4/zephyr-7b-alpha~\cite{arxiv:2305.14233,arxiv:2310.16944,arxiv:2310.01377,arxiv:2305.18290}} & mistralai/Mistral-7B-v0.1 & mistral & 2 & 12911 \\
55 & \parbox[t]{6cm}{HuggingFaceH4/zephyr-7b-beta~\cite{arxiv:2305.14233,arxiv:2310.16944,arxiv:2310.01377,arxiv:2305.18290}} & mistralai/Mistral-7B-v0.1 & mistral & 2 & 294019 \\
56 & \parbox[t]{6cm}{ibm/merlinite-7b~\cite{arxiv:2403.01081}} & mistralai/Mistral-7B-v0.1 & mistral & 2 & 11156 \\
57 & \parbox[t]{6cm}{INSAIT-Institute/BgGPT-7B-Instruct-v0.2} & mistralai/Mistral-7B-v0.1 & mistral & 2 & 3265 \\
58 & \parbox[t]{6cm}{Intel/neural-chat-7b-v3-1~\cite{arxiv:2306.02707}} & mistralai/Mistral-7B-v0.1 & mistral & 2 & 3976 \\
59 & \parbox[t]{6cm}{kaist-ai/mistral-orpo-capybara-7k~\cite{arxiv:2403.07691}} & mistralai/Mistral-7B-v0.1 & mistral & 2 & 4821 \\
60 & \parbox[t]{6cm}{Locutusque/Hercules-2.5-Mistral-7B} & mistralai/Mistral-7B-v0.1 & mistral & 2 & 1953 \\
61 & \parbox[t]{6cm}{mistralai/Mistral-7B-Instruct-v0.1~\cite{arxiv:2310.06825}} & mistralai/Mistral-7B-v0.1 & mistral & 2 & 1370245 \\
62 & \parbox[t]{6cm}{NousResearch/Hermes-2-Pro-Mistral-7B~\cite{Hermes-2-Pro-Mistral-7B}} & mistralai/Mistral-7B-v0.1 & mistral & 2 & 14586 \\
63 & \parbox[t]{6cm}{NousResearch/Nous-Hermes-2-Mistral-7B-DPO~\cite{Nous-Hermes-2-Mistral-7B-DPO}} & mistralai/Mistral-7B-v0.1 & mistral & 2 & 8725 \\
64 & \parbox[t]{6cm}{openchat/openchat-3.5-0106~\cite{arxiv:2309.11235,arxiv:2303.08774}} & mistralai/Mistral-7B-v0.1 & mistral & 2 & 26858 \\
65 & \parbox[t]{6cm}{openchat/openchat-3.5-1210~\cite{arxiv:2309.11235,arxiv:2303.08774}} & mistralai/Mistral-7B-v0.1 & mistral & 2 & 2123 \\
66 & \parbox[t]{6cm}{statking/zephyr-7b-sft-full-orpo} & mistralai/Mistral-7B-v0.1 & mistral & 2 & 2278 \\
67 & \parbox[t]{6cm}{teknium/OpenHermes-2.5-Mistral-7B} & mistralai/Mistral-7B-v0.1 & mistral & 2 & 100997 \\
68 & \parbox[t]{6cm}{mlabonne/NeuralMarcoro14-7B} & mlabonne/Marcoro14-7B-slerp & mistral & 2 & 2015 \\
69 & \parbox[t]{6cm}{mlabonne/AlphaMonarch-7B} & mlabonne/NeuralMonarch-7B & mistral & 2 & 12804 \\
70 & \parbox[t]{6cm}{augmxnt/shisa-gamma-7b-v1} & stabilityai/japanese-stablelm-base-gamma-7b & mistral & 2 & 151426 \\
71 & \parbox[t]{6cm}{beomi/KoAlpaca-KoRWKV-6B} & beomi/KoRWKV-6B & others & 2 & 2288 \\
72 & \parbox[t]{6cm}{Nexusflow/NexusRaven-V2-13B~\cite{nexusraven,arxiv:2308.12950}} & codellama/CodeLlama-13b-Instruct-hf & others & 2 & 3966 \\
73 & \parbox[t]{6cm}{haoranxu/ALMA-13B-Pretrain~\cite{arxiv:2309.11674,arxiv:2401.08417}} & meta-llama/Llama-2-13b-hf & others & 2 & 5510 \\
74 & \parbox[t]{6cm}{ruslanmv/ai-medical-model-32bit} & meta-llama/Meta-Llama-3-8B-Instruct & others & 2 & 2810 \\
75 & \parbox[t]{6cm}{Locutusque/Orca-2-13b-SFT-v4} & microsoft/Orca-2-13b & others & 2 & 2345 \\
76 & \parbox[t]{6cm}{Locutusque/Orca-2-13b-SFT-v6} & microsoft/Orca-2-13b & others & 2 & 2319 \\
77 & \parbox[t]{6cm}{Locutusque/Orca-2-13b-SFT\_v5} & microsoft/Orca-2-13b & others & 2 & 2191 \\
78 & \parbox[t]{6cm}{mwitiderrick/open\_llama\_3b\_code\_instruct\_0.1} & openlm-research/open\_llama\_3b & others & 2 & 1252 \\
79 & \parbox[t]{6cm}{NousResearch/Nous-Hermes-2-SOLAR-10.7B} & upstage/SOLAR-10.7B-v1.0 & others & 2 & 9918 \\
80 & \parbox[t]{6cm}{upstage/SOLAR-10.7B-Instruct-v1.0~\cite{arxiv:2312.15166,arxiv:2403.19270}} & upstage/SOLAR-10.7B-v1.0 & others & 2 & 67725 \\
81 & \parbox[t]{6cm}{cognitivecomputations/dolphin-2.9.1-yi-1.5-9b} & 01-ai/Yi-1.5-9B & others & 2 & 4880 \\
82 & \parbox[t]{6cm}{deepseek-ai/DeepSeek-Prover-V1.5-RL~\cite{arxiv:2408.08152}} & deepseek-ai/DeepSeek-Prover-V1.5-SFT & deepseek & 3 & 12223 \\
83 & \parbox[t]{6cm}{euclaise/falcon\_1b\_stage2} & euclaise/falcon\_1b\_stage1 & falcon & 3 & 4016 \\
84 & \parbox[t]{6cm}{MaziyarPanahi/Llama-3-8B-Instruct-v0.8} & MaziyarPanahi/Llama-3-8B-Instruct-v0.4 & llama-3 & 3 & 7281 \\
85 & \parbox[t]{6cm}{argilla/notus-7b-v1} & alignment-handbook/zephyr-7b-sft-full & mistral & 3 & 7660 \\
86 & \parbox[t]{6cm}{Intel/neural-chat-7b-v3-3~\cite{arxiv:2309.12284}} & Intel/neural-chat-7b-v3-1 & mistral & 3 & 166226 \\
87 & \parbox[t]{6cm}{MaziyarPanahi/Llama-3-8B-Instruct-v0.9} & MaziyarPanahi/Llama-3-8B-Instruct-v0.8 & llama-3 & 4 & 6241 \\

\bottomrule
\label{tab:model_list_ft}
\end{longtable}
}

{\scriptsize
\begin{longtable}{r@{\hspace{1em}}l@{\hspace{1em}}l@{\hspace{1em}}r@{\hspace{1em}}r}
\caption{List of 50 (+9) models in the experiments for quantization and layer-wise models. ``ID'' denotes the index sorted in descending order of downloads; ``Model Name'' denotes the name of the model; ``Model Type'' denotes the model classification defined in \citet{DBLP:conf/acl/OyamaYTS25}; ``Layers'' denotes the number of layers in the model; ``DLs'' denotes the total number of downloads.}\\

\toprule
ID & Model Name & Model Type & Layers & DLs\\
\midrule
\endfirsthead
    
\toprule
ID & Model Name & Model Type & Layers & DLs\\
\midrule
\endhead
1 & \parbox[t]{11.25cm}{mistralai/Mistral-7B-Instruct-v0.2~\cite{arxiv:2310.06825}} & mistral & 32 & 3565248 \\
2 & \parbox[t]{11.25cm}{meta-llama/Meta-Llama-3-8B-Instruct~\cite{llama3modelcard}} & llama-3 & 32 & 2058011 \\
3 & \parbox[t]{11.25cm}{mistralai/Mistral-7B-v0.1~\cite{arxiv:2310.06825}} & mistral & 32 & 1750089 \\
4 & \parbox[t]{11.25cm}{mistralai/Mistral-7B-v0.3} & mistral & 32 & 1443065 \\
5 & \parbox[t]{11.25cm}{meta-llama/Llama-2-7b-chat-hf~\cite{arxiv:2307.09288}} & llama-2 & 32 & 1402244 \\
6 & \parbox[t]{11.25cm}{mistralai/Mistral-7B-Instruct-v0.1~\cite{arxiv:2310.06825}} & mistral & 32 & 1370245 \\
7 & \parbox[t]{11.25cm}{meta-llama/Llama-2-7b-hf~\cite{arxiv:2307.09288}} & llama-2 & 32 & 1294737 \\
8 & \parbox[t]{11.25cm}{TinyLlama/TinyLlama-1.1B-Chat-v1.0} & others & 22 & 1078500 \\
9 & \parbox[t]{11.25cm}{TinyLlama/TinyLlama-1.1B-intermediate-step-1431k-3T} & others & 22 & 798206 \\
10 & \parbox[t]{11.25cm}{meta-llama/Meta-Llama-3-8B~\cite{llama3modelcard}} & llama-3 & 32 & 675850 \\
11 & \parbox[t]{11.25cm}{OpenAssistant/oasst-sft-4-pythia-12b-epoch-3.5} & gpt\_neox & 36 & 458718 \\
12 & \parbox[t]{11.25cm}{bigcode/starcoder2-3b~\cite{arxiv:2004.05150,arxiv:2205.14135,arxiv:2207.14255,arxiv:2305.13245,arxiv:2402.19173}} & others & 30 & 445316 \\
13 & \parbox[t]{11.25cm}{lmsys/vicuna-7b-v1.5~\cite{arxiv:2307.09288,arxiv:2306.05685}} & llama-2 & 32 & 368410 \\
14 & \parbox[t]{11.25cm}{HuggingFaceH4/zephyr-7b-beta~\cite{arxiv:2305.14233,arxiv:2310.16944,arxiv:2310.01377,arxiv:2305.18290}} & mistral & 32 & 294019 \\
15 & \parbox[t]{11.25cm}{meta-llama/Llama-2-13b-chat-hf~\cite{arxiv:2307.09288}} & llama-2 & 40 & 266090 \\
16 & \parbox[t]{11.25cm}{google/gemma-2b~\cite{arxiv:1705.03551,arxiv:1809.02789,arxiv:1804.06876,arxiv:1804.09301,arxiv:1911.11641,arxiv:1911.01547,arxiv:1905.10044,arxiv:1811.00937,arxiv:1904.09728,arxiv:1905.07830,arxiv:1907.10641,arxiv:2009.11462,arxiv:2107.03374,arxiv:2108.07732,arxiv:2009.03300,arxiv:2101.11718,arxiv:2110.14168,arxiv:2109.07958,arxiv:2203.09509,arxiv:2110.08193,arxiv:2206.04615,arxiv:2304.06364,arxiv:2312.11805}} & gemma & 18 & 256798 \\
17 & \parbox[t]{11.25cm}{EleutherAI/gpt-neo-1.3B~\cite{arxiv:2101.00027,gpt-neo}} & others & 24 & 243332 \\
18 & \parbox[t]{11.25cm}{EleutherAI/gpt-j-6b~\cite{arxiv:2101.00027,gpt-j,mesh-transformer-jax,arxiv:2104.09864}} & gptj & 28 & 241435 \\
19 & \parbox[t]{11.25cm}{microsoft/phi-2} & others & 32 & 237268 \\
20 & \parbox[t]{11.25cm}{EleutherAI/gpt-neo-2.7B~\cite{arxiv:2101.00027,gpt-neo}} & others & 32 & 205503 \\
21 & \parbox[t]{11.25cm}{huggyllama/llama-7b} & llama-1 & 32 & 192383 \\
22 & \parbox[t]{11.25cm}{tiiuae/falcon-7b-instruct~\cite{arxiv:1911.02150,arxiv:2005.14165,arxiv:2205.14135,arxiv:2104.09864,falcon40b,arxiv:2306.01116}} & falcon & 32 & 179952 \\
23 & \parbox[t]{11.25cm}{DeepMount00/Llama-3-8b-Ita} & llama-3 & 32 & 179401 \\
24 & \parbox[t]{11.25cm}{Intel/neural-chat-7b-v3-3~\cite{arxiv:2309.12284}} & mistral & 32 & 166226 \\
25 & \parbox[t]{11.25cm}{openlm-research/open\_llama\_3b~\cite{together2023redpajama,arxiv:2302.13971,openlm2023openllama}} & llama-1 & 26 & 157405 \\
26 & \parbox[t]{11.25cm}{augmxnt/shisa-gamma-7b-v1} & mistral & 32 & 151426 \\
27 & \parbox[t]{11.25cm}{facebook/opt-1.3b~\cite{arxiv:2005.14165,arxiv:2205.01068}} & opt & 24 & 139758 \\
28 & \parbox[t]{11.25cm}{Qwen/Qwen1.5-1.8B~\cite{arxiv:2309.16609}} & others & 24 & 137256 \\
29 & \parbox[t]{11.25cm}{codellama/CodeLlama-7b-Instruct-hf~\cite{arxiv:2308.12950}} & llama-2 & 32 & 133523 \\
30 & \parbox[t]{11.25cm}{cognitivecomputations/dolphin-2.9-llama3-8b} & llama-3 & 32 & 130977 \\
31 & \parbox[t]{11.25cm}{Qwen/Qwen1.5-7B~\cite{arxiv:2309.16609}} & others & 32 & 122768 \\
32 & \parbox[t]{11.25cm}{meta-llama/Llama-2-13b-hf~\cite{arxiv:2307.09288}} & llama-2 & 40 & 120871 \\
33 & \parbox[t]{11.25cm}{microsoft/phi-1\_5~\cite{arxiv:2309.05463}} & others & 24 & 112476 \\
34 & \parbox[t]{11.25cm}{NousResearch/Meta-Llama-3-8B-Instruct~\cite{llama3modelcard}} & llama-3 & 32 & 104388 \\
35 & \parbox[t]{11.25cm}{tiiuae/falcon-7b~\cite{arxiv:1911.02150,arxiv:2005.14165,arxiv:2101.00027,arxiv:2205.14135,arxiv:2104.09864,falcon40b,arxiv:2306.01116}} & falcon & 32 & 104010 \\
36 & \parbox[t]{11.25cm}{google/gemma-2b-it~\cite{arxiv:1705.03551,arxiv:1809.02789,arxiv:1804.06876,arxiv:1804.09301,arxiv:1911.11641,arxiv:1911.01547,arxiv:1905.10044,arxiv:1811.00937,arxiv:1904.09728,arxiv:1905.07830,arxiv:1907.10641,arxiv:2009.11462,arxiv:2107.03374,arxiv:2108.07732,arxiv:2009.03300,arxiv:2101.11718,arxiv:2110.14168,arxiv:2109.07958,arxiv:2203.09509,arxiv:2110.08193,arxiv:2206.04615,arxiv:2304.06364,arxiv:2312.11805}} & gemma & 18 & 103851 \\
37 & \parbox[t]{11.25cm}{teknium/OpenHermes-2.5-Mistral-7B} & mistral & 32 & 100997 \\
38 & \parbox[t]{11.25cm}{mosaicml/mpt-7b-chat~\cite{arxiv:2010.04245,arxiv:2108.12409,arxiv:2205.14135,MosaicML2023Introducing}} & others & 32 & 88069 \\
39 & \parbox[t]{11.25cm}{Qwen/CodeQwen1.5-7B-Chat~\cite{arxiv:2309.16609}} & others & 32 & 77169 \\
40 & \parbox[t]{11.25cm}{GritLM/GritLM-7B~\cite{arxiv:2402.09906}} & mistral & 32 & 72991 \\
41 & \parbox[t]{11.25cm}{google/gemma-7b~\cite{arxiv:1705.03551,arxiv:1809.02789,arxiv:1804.06876,arxiv:1804.09301,arxiv:1911.11641,arxiv:1911.01547,arxiv:1905.10044,arxiv:1811.00937,arxiv:1904.09728,arxiv:1905.07830,arxiv:1907.10641,arxiv:2009.11462,arxiv:2107.03374,arxiv:2108.07732,arxiv:2009.03300,arxiv:2101.11718,arxiv:2110.14168,arxiv:2109.07958,arxiv:2203.09509,arxiv:2110.08193,arxiv:2206.04615,arxiv:2304.06364,arxiv:2305.14314,arxiv:2312.11805}} & gemma & 28 & 72047 \\
42 & \parbox[t]{11.25cm}{upstage/SOLAR-10.7B-Instruct-v1.0~\cite{arxiv:2312.15166,arxiv:2403.19270}} & others & 48 & 67725 \\
43 & \parbox[t]{11.25cm}{google/gemma-7b-it~\cite{arxiv:1705.03551,arxiv:1809.02789,arxiv:1804.06876,arxiv:1804.09301,arxiv:1911.11641,arxiv:1911.01547,arxiv:1905.10044,arxiv:1811.00937,arxiv:1904.09728,arxiv:1905.07830,arxiv:1907.10641,arxiv:2009.11462,arxiv:2107.03374,arxiv:2108.07732,arxiv:2009.03300,arxiv:2101.11718,arxiv:2110.14168,arxiv:2109.07958,arxiv:2203.09509,arxiv:2110.08193,arxiv:2206.04615,arxiv:2304.06364,arxiv:2312.11805}} & gemma & 28 & 66776 \\
44 & \parbox[t]{11.25cm}{codellama/CodeLlama-7b-hf~\cite{arxiv:2308.12950}} & llama-2 & 32 & 66040 \\
45 & \parbox[t]{11.25cm}{facebook/opt-2.7b~\cite{arxiv:2005.14165,arxiv:2205.01068}} & opt & 32 & 61450 \\
46 & \parbox[t]{11.25cm}{shenzhi-wang/Llama3-8B-Chinese-Chat~\cite{shenzhi-wang-2024}} & llama-3 & 32 & 55718 \\
47 & \parbox[t]{11.25cm}{VAGOsolutions/Llama-3-SauerkrautLM-8b-Instruct} & llama-3 & 32 & 55400 \\
48 & \parbox[t]{11.25cm}{lmsys/vicuna-13b-v1.5~\cite{arxiv:2307.09288,arxiv:2306.05685}} & llama-2 & 40 & 48653 \\
49 & \parbox[t]{11.25cm}{Qwen/Qwen2-1.5B~\cite{qwen2}} & others & 28 & 46510 \\
50 & \parbox[t]{11.25cm}{openlm-research/open\_llama\_7b~\cite{together2023redpajama,arxiv:2302.13971,openlm2023openllama}} & llama-1 & 32 & 44968 \\
51 & \parbox[t]{11.25cm}{LinkSoul/Chinese-Llama-2-7b} & llama-2 & 32 & 40799 \\
52 & \parbox[t]{11.25cm}{mistral-community/Mistral-7B-v0.2} & mistral & 32 & 30074 \\
53 & \parbox[t]{11.25cm}{NousResearch/Hermes-2-Pro-Llama-3-8B~\cite{Hermes-2-Pro-Llama-3-8B}} & llama-3 & 32 & 26797 \\
54 & \parbox[t]{11.25cm}{lightblue/suzume-llama-3-8B-multilingual~\cite{arxiv:2405.12612}} & llama-3 & 32 & 16442 \\
55 & \parbox[t]{11.25cm}{abacusai/Llama-3-Smaug-8B~\cite{arxiv:2402.13228}} & llama-3 & 32 & 12873 \\
56 & \parbox[t]{11.25cm}{NousResearch/Nous-Hermes-llama-2-7b} & llama-2 & 32 & 12019 \\
57 & \parbox[t]{11.25cm}{togethercomputer/LLaMA-2-7B-32K} & llama-2 & 32 & 8884 \\
58 & \parbox[t]{11.25cm}{elyza/ELYZA-japanese-Llama-2-7b-instruct~\cite{arxiv:2307.09288,elyzallama2023}} & llama-2 & 32 & 6296 \\
59 & \parbox[t]{11.25cm}{togethercomputer/Llama-2-7B-32K-Instruct~\cite{arxiv:2307.03172}} & llama-2 & 32 & 5596 \\

\bottomrule
\label{tab:model_list_layer}
\end{longtable}
}

\end{document}